\documentclass[english]{IEEEtran}
\usepackage[T1]{fontenc}
\usepackage[latin9]{inputenc}
\usepackage{prettyref}
\usepackage{float}
\usepackage{amsmath}
\usepackage{amssymb}
\usepackage{graphicx}
\usepackage{esint}

\makeatletter

\floatstyle{ruled}
\newfloat{algorithm}{tbp}{loa}
\providecommand{\algorithmname}{Algorithm}
\floatname{algorithm}{\protect\algorithmname}

\usepackage{algorithm}\usepackage{algpseudocode}
\usepackage{babel}

\makeatother

\usepackage{babel}
\begin{document}

\title{Particle Smoothing for Hidden Diffusion Processes:\\
 Adaptive Path Integral Smoother}

\author{H.-Ch. Ruiz and H.J. Kappen \thanks{The work of H.-Ch. Ruiz is supported by the European Commission through the FP7 Marie Curie Initial Training Network 289146, NETT: Neural Engineering Transformative Technologies.} \thanks{The authors are with the Biophysics Department at Donders Institute for Brain, Cognition and Behaviour, Radboud University, 6525AJ Nijmegen, The Netherlands (email: H.Ruiz@science.ru.nl; B.Kappen@science.ru.nl).}}
\maketitle
\begin{abstract}
Smoothing methods are used for inference of stochastic processes given
noisy observations. The estimation of the marginal posterior distribution
given all observations is typically a computationally intensive task.
We propose a novel algorithm based on path integral control theory
to efficiently estimate the smoothing distribution of continuous-time
diffusion processes from partial observations. In particular, we use
an adaptive importance sampling method to improve the effective sampling
size of the posterior and the reliability of the estimation of the
marginals. This is achieved by estimating a feedback controller to
help sample efficiently from the joint smoothing distribution. We
compare the results with estimations obtained from the standard Forward
Filter/Backward Simulator (FFBSi) for two diffusion processes of different
complexity. We show that the proposed method gives more accurate estimates
than the standard FFBSi. 
\end{abstract}

\section{Introduction}

\subsection*{Problem Statement}

In many fields of science and engineering access to physical time
varying processes is limited to time series of noisy, indirect measurements.
In order to extract information about the latent process, one estimates
the so-called filtering or smoothing distributions. It is then possible
to estimate the time evolution of the latent states, or estimate the
parameters of a model, for example using an Expectation-Maximization
procedure.

In this paper, we consider the smoothing problem for continuous time
diffusion processes given a discrete number of observations. The latent
process $X_{t}$ is described by the following $n$-dimensional stochastic
differential equation (SDE) 
\begin{equation}
dX_{t}=F(X_{t},t)dt+\sigma_{dyn}(X_{t},t)dW_{t}\label{eq:Euler-Maruyama}
\end{equation}
where $dW_{t}$ is a $m$-dimensional Gaussian noise with $\mathbb{E}\left[dW_{t}\right]=0$
and $\mathbb{E}\left[dW_{r}^{i}dW_{s}^{j}\right]=dt\delta_{i,j}\delta(r-s)$
and $\sigma_{dyn}(x,t)\in\mathbb{R}^{n\times m}$ is a matrix that
depends on the state $x$ and time $t$. For given initial state $x_{0}$,
(\ref{eq:Euler-Maruyama}) defines a distribution over processes $p_{0}(X_{(0,T]}|x_{0})$.
When the initial state $x_{0}$ is drawn from a distribution $p_{0}(X_{0})$,
this defines a prior distribution over processes $p_{0}(X_{[0,T]})=p_{0}(X_{0})p_{0}(X_{(0,T]}|X_{0})$.

We assume an observation model $g(y|x)$ that denotes the probability
of observation $y$ at time $t$ given the latent state $x$ at time
$t$. Given $J$ observations $y_{t_{j}}$ at times $t_{j}$, with
$t_{j}\in[0,T]$ for all $j=1,\dots,J$ and $t_{J}=T$, this defines
a likelihood $p(y_{0:T}|X_{[0,T]})=\prod_{j=1}^{J}g(y_{t_{j}}|X_{t_{j}})$.
The smoothing problem is to estimate marginals or statistics of the
posterior distribution, also referred to as the smoothing distribution:
\begin{equation}
p(X_{[0,T]}|y_{0:T})=\frac{1}{Z}p_{0}(X_{[0,T]})\exp\left(\sum_{j=1}^{J}\log[g(y_{t_{j}}|X_{t_{j}})]\right).\label{eq:smoothing-distribution}
\end{equation}
with $Z=p(y_{0:T})$ the likelihood of the data%
\footnote{We denote time series of discrete observations by $y_{0:T}:=(y_{t_{1}},y_{t_{2}},\dots,y_{t_{J}})$
and continuous paths by $X_{[0,T]}:=(X_{s})_{s\in[0,T]\subset\mathbb{R}}$.%
}.

The smoothing problem is in general intractable when the dynamics
(\ref{eq:Euler-Maruyama}) is non-linear or when the observation model
is non-Gaussian. In those cases, it is needed to resort to approximate
methods. One class of these methods is the deterministic approximation
methods such as non-linear Kalman filtering \cite{JulierUhlmann1997,sarkka2007unscented}
and smoothing \cite{sarkka2008unscented}, or the variational method
\cite{sutter2015variational}, which approximate the posterior by
a simpler distribution. These methods are relatively efficient but
may be inaccurate in some cases and will not be considered further
in this paper.

In the remaining of this section, we will discuss three alternative
classes of smoothing methods, first, particle filtering, second, adaptive
importance sampling, and third, inference as a control problem. In
the latter class, we will introduce our method 'Adaptive Path Integral
Smoother'.

\subsection{Particle Filtering Methods}

A prominent sampling based method, known as Sequential Monte Carlo
(SMC) sampling or particle filtering is used to target the smoothing
distribution. Particle filtering methods estimate the smoothing distribution
by computing estimates of the filtered distribution and subsequently
correct for these estimates. Each particle corresponds to an entire
trajectory $X_{[0,T]}$. Among the various SMC methods for smoothing,
one can distinguish broadly speaking three approaches; first, the
bootstrap Filter-Smoother (FS) by \cite{kitagawa1996monte}, second,
the forward-backward smoothers \cite{Godsill2004,Doucet2000}--with
its many variations \cite{Lindsten2013}--and third, the two-filter
smoothers \cite{bresler1986two,Briers2010,Fearnhead2010}. All these
methods have their particular strengths and weaknesses. See e.g. \cite{Doucet2009,liu2008monte,fearnhead2008computational}
for a review on various filtering methods.

In naive particle smoothing each particle is sampled from forward
simulation of $p_{0}(X_{[0,T]})$ and weighted with $w=\exp\left(\sum_{j=1}^{J}\log[g(y_{t_{j}}|X_{t_{j}})]\right)$.
With many observations (large $J$), the so-called degeneracy problem
is introduced, where the weight $w$ of one particle dominates all
other weights. As a result, the representation of the smoothing distribution
is very poor.

One can reduce the degeneracy by resampling the filtering particles.
In its simplest form, the resampling step is done at each observation,
but more sophisticated adaptive schemes exist \cite{del2012adaptive}.
Resampling is an effective way to improve the quality of the filtered
estimates. 

The trajectories of the resampled particles can also be used to estimate
the smoothing distribution, as in the bootstrap Filter-Smoother (FS)
\cite{kitagawa1996monte}. However, the effect of resampling is that
all trajectories arise from a very small number of common trajectories
at early times. As a result of this \textquotedbl{}path degeneracy\textquotedbl{},
the resampled trajectories give a poor representation of the smoothed
marginals $p(X_{t}|y_{0:T})$ at early times $t\ll T$. The path degeneracy
increases also exponentially fast as $J$ increases \cite{Chopin2004}.
In other words, resampling improves the filtered estimates but not
the smoothed estimates.

The degeneracy problem is particularly severe when the observations
deviate significantly from the prior process. In this case, the smoothing
distribution may be very different from the filtering distribution
causing weights with high variance and low effective sample size.
As a result, the number of particles $N$ needs to be prohibitively
large to have moderate accuracy. 

The quality of the smoothing estimates can be improved by adding a
backward simulation, known as Forward Filter Backward Simulator (FFBSi)
\cite{Godsill2004} which obtains trajectories approximately from
the joint posterior. Applying the backward pass with $M$ particles
has a complexity ${\cal O}(MN)$. Since typically $M={\cal O}(N)$
backward particles are required, the accuracy of this method is severely
limited in practice by the computational cost. Several approaches
have been developed to lower the computational effort while maintaining
reliable estimates. For instance, in \cite{Douc2011} a rejection
sampling approach was suggested to avoid the computational complexity
of evaluating all backward weights, effectively reducing the overall
computational complexity to ${\cal O}(N)$ provided that $N$ is sufficiently
large. However in practice, this approach is less efficient than FFBSi
for many problems and does not scale to high dimensions \cite{Lindsten2013}.

The Forward Filter Backward Smoother \cite{Doucet2000} aims at approximating
the marginal smoothing densities. This is done by reweighting the
forward filter particles to target the posterior marginals. The computational
complexity is ${\cal O}(N^{2})$ due to the reweighting step.

An additional limitation of the backward methods, aside from their
computational demands, is that they assume the existence of a non-degenerate
backward kernel. In the case of the process (\ref{eq:Euler-Maruyama}),
this means that the noise covariance matrix $\sigma_{dyn}\sigma_{dyn}'$
must be non-singular, which limits the applicability, for instance
when the dynamics of some components of $X_{t}$ is deterministic.

Finally, the forward-backward approaches have a further limitation
in continuous time problems. The efficiency of the integration of
SDEs can be increased significantly by replacing the standard Euler-Maruyama
integration by a higher order scheme \cite{Kloeden2012}. Since higher
order schemes cannot be used in the backward computation step, the
overall efficiency of backward methods can not be improved by these
integration methods. See, however, \cite{Murray2011} for some interesting
work that allows higher order integration schemes using kernel density
estimation.

Another approach within the SMC methods is the generalized two-filter
smoother \cite{Briers2010}, which involves sampling from both a backward
information filter and a forward particle filter. The particles of
both filters are combined to obtain an approximation of the marginal
$p(x_{t}|y_{0:T})$, which is used to sample approximately from the
joint smoothing distribution. The method requires the choice of an
artificial prior at each time point affecting the efficiency of the
sampler. Besides, this method also requires ${\cal O}(N^{2})$ samples.

As noted by \cite{Fearnhead2010}, whenever the forward state transition
probability $f(X_{t}|X_{t-dt})$ is approximately zero for most state
pairs $X_{t-dt}^{j},X_{t}^{i}$ (sparse dynamics), the forward-backward
smoother degenerates to being equivalent to the filter-smoother, albeit
with substantially greater computational cost. The situation is worse
for the two-filter smoother which fails completely as the forward
and backward filter particles are sampled independently. This problem
is particularly relevant for continuous time stochastic systems. Here,
the variance of $dX_{t}$ is proportional to $dt$ thus, the transition
probability from particle $j$ at time $t-dt$ to particle $i$ at
time $t$ is exponentially suppressed for all pairs $i\ne j$.

This issue is addressed in \cite{Fearnhead2010} by drawing new particles
from the smoothing marginals directly.\textbf{ }Although, the computational
complexity of this approach is linear in the number of particles,
it is not clear how to choose the required artificial densities in
general. As a result, the method suffers from cumbersome design choices
\cite{Taghavi2012} which makes it impractical in many cases.

Other approaches that can ameliorate the particle degeneracy are developed
in \cite{Dubarry2011,Bunch2013}. Both methods propose to use Metropolis-Hastings
moves to sample new positions and generate trajectories of the joint
smoothing distribution given an existing particle system. In principle,
this could move particles to higher density regions of the smoothing
distribution and increase the effective sample size. In \cite{Dubarry2011},
the Metropolis-Hastings Improved Particle Smoother (MH-IPS) uses Gibbs
sampling to sample a new state $X_{t}$ given the remaining particle
states. However, this method might be subject to strong dependencies
between state variables, resulting in a poor mixing whenever the discretization
time $dt$ of the underlying SDE is sufficiently small.

Recently, \cite{guarniero2015iterated} considered so called twisted
models based on the idea of message passing through the Markov representation
of the posterior Eq.~(\ref{eq:smoothing-distribution}). The messages
are positive functions that need to be approximated iteratively. This
is done by sampling from a \textquotedbl{}twisted\textquotedbl{} auxiliary
particle filter and using the particles to estimate new messages.
The disadvantage of this method is that in practice the transition
density and messages are restricted to certain classes. 

All the above methods have a particle filtering step in common.

\subsection{Adaptive Importance Sampling}

Importance sampling is a way of obtaining samples from a target distribution
indirectly. The idea is to sample from a proposal distribution that
is different from the target distribution and to weight the samples
by importance sampling weights. Adaptive importance sampling \cite{de2005tutorial,cappe2008adaptive}
adapts the parameters of the proposal distribution by minimizing some
cost criterion, such as the Kullback-Leibler (KL) divergence or the
chi-square distance between proposal and target distributions.

In \cite{cornebise2008adaptive} an adaptive importance sampling method
is proposed for time-series models. This work uses an auxiliary particle
filter \cite{pitt1999filtering} to construct adjustment multiplier
weights that minimize the aforementioned risk criteria for a given
proposal kernel. In addition, optimization techniques are proposed
to adjust the proposal kernels by minimizing the risk criteria. For
instance, the KL divergence is minimized using the cross-entropy method.
To the best of our knowledge, this method has not been applied to
the continuous time smoothing problem.

\subsection{Inference as a Control Problem}

A fundamentally different approach to address the smoothing and the
degeneracy problem is to 'steer' the particles through time based
on future observations. Steering is optimal when the degeneracy problem
is solved. In this sense, the smoothing problem can be viewed as a
stochastic optimal control problem. The relationship between control
and inference was first established by \cite{Pardoux1981,Fleming1982,Mitter1982}
who showed that the posterior inference for the smoothing problem
(\ref{eq:smoothing-distribution}) can be mapped onto a certain class
of so called path integral control problems. In \cite{Kappen2005},
it was shown how to compute the optimal control for these problems.
Thus far, few authors have considered the application of this idea
for smoothing. In this paper, we propose such an algorithm.

Nevertheless, we briefly review other approaches to inference that
use ideas from control theory, but not from within the path integral
control theory. In \cite{yang2013feedback}, it is shown that for
a general non-linear diffusion with non-Gaussian observations, the
optimal (state-dependent) Kalman gain can be computed at each time
as an Euler-Lagrange boundary value problem. However, the approach
is restricted to one-dimensional diffusion processes only. In \cite{pequito2011nonlinear},
it is proposed to improve the posterior estimate by considering interacting
particles. These so-called mean field game systems describe interacting
particles whose density evolves according to a (forward) Fokker-Planck
equation which is controlled by a (backward) Hamilton-Jacobi-Bellman
equation. The disadvantage of this approach is that one needs to solve
the HJB equation which is intractable for high dimensions.

In \cite{Fleming1982}, the authors showed that the smoothing distribution
Eq.~(\ref{eq:smoothing-distribution}) can be sampled with Eq.~(\ref{eq:controled process}),
which differs from (\ref{eq:Euler-Maruyama}) by a control term $u(x,t)$.
The function $u(x,t)$ must be chosen optimally to minimize a control
cost. The optimal control can be estimated for each $x,t$ as a path
integral. It can be shown that the optimal control gives the optimal
(zero variance) importance sampler. In general, we cannot compute
the optimal control function for all $x,t$. For the smoothing problem,
we therefore propose a parametrized controller and learn the parameters
by an iterative scheme, that was first proposed in \cite{Thijssen2015}.
We call this method Adaptive Path Integral Smoother (APIS). APIS iteratively
reduces the variance of the weights for a given time-series and thus
improves the sampling efficiency in terms of effective sample size.
This improvement is limited mainly by the class of control functions
that is considered. If the correct parametrization of the optimal
control solution is available, the effective sample size is only limited
by the numerical errors coming from the time discretization and the
sample error. As a result, APIS requires increasing precision to maintain
the sampling efficiency for longer time series, i.e. more particles
and smaller integration steps are needed. In this paper, we restrict
ourselves to linear state-feedback controllers and we show that these
yield very reliable smoothing estimates even when used in non-linear
systems.

An additional advantage of APIS for continuous time problems is that
it does not contain a backward step, so it can be accelerated by using
higher order integration schemes. Furthermore, there is not restriction
on the degeneracy of the covariance matrix $\sigma_{dyn}\sigma_{dyn}'$.
This is particularly useful for problems with mixed deterministic
and stochastic dynamics. Finally, the variance of the estimates are
not increased due to resampling because APIS does not require this
step \cite{Doucet2009,Chopin2004}.

In \cite{kappen2016adaptive}, preliminary results were shown on a
small problem. In this paper, we provide the full detailed description
of the implementation of the APIS method, and extend the method with
a novel adaptive initialization of the particles and a novel annealing/bootstrapping
scheme, which are both crucial for the sampling efficiency, in particular
for large time series with many observations. In addition, we analyze
in detail the quality of APIS in terms of effective sample size, we
compare APIS with the vanilla flavor FFBSi and FS particle filtering
algorithms and we analyze the scalability of APIS for up to 1000 observations.

\subsection*{Outline}

This paper is organized as follows. In section~\ref{sec:PIcontrol}
we review the main concepts in path integral (PI) control theory.
We show how computing the joint smoothing distribution in continuous
time is equivalent to a PI control problem. In section~\ref{sec:Method},
we discuss the importance sampling scheme for diffusion processes
based on control. Then, we give an update rule to estimate a feedback
controller and present the APIS algorithm. In section~\ref{sec:Results}
we present numerical examples. First, we consider the simple case
of a one-dimensional linear diffusion process with Gaussian observations.
We compare the accuracy and efficiently of FFBSi, the Bootstrap Filter-Smoother
(FS) and APIS and show their performance as a function of the (un)likelihood
of the observations. In addition, we examine the scaling of APIS up
to 1000 observations. Then, we consider a 5-dimensional non-linear
neural network model with multiple Gaussian observations and show
that even a suboptimal linear feedback controller improves drastically
the ESS. Moreover, we show that the estimation of the smoothing distribution
is more reliable with APIS. In section~\ref{sec:Discussion} we comment
on further considerations for the proposed algorithm. Finally, we
outline possible extensions of this method that will be addressed
in future work.

\section{PI Control Theory and the Smoothing Distribution\label{sec:PIcontrol}}

We introduce the basic concepts regarding a subclass of stochastic
control problems called Path Integral control problems, for more details
see \cite{Kappen2005,Kappen2011,Kappen2012}.

Stochastic optimal control theory considers systems under uncertain
time evolution. The aim is to compute the optimal feedback control
function to steer the system to a specified future goal. More formally,
we have a continuous time stochastic process $X_{t}$ ($t\in[0,T]$)
described by the following $n$-dimensional SDE with the initial condition
$X_{0}=x_{0}$ 
\begin{equation}
dX_{t}=F(X_{t},t)dt+\sigma_{dyn}(X_{t},t)[u(X_{t},t)dt+dW_{t}]\label{eq:controled process}
\end{equation}
where $dW_{t}$ and $\sigma_{dyn}(x,t)$ are as before in (\ref{eq:Euler-Maruyama}).
We denote%
\footnote{Note that we also distinguish between a deterministic function of
state and time, e.g. $\sigma_{dyn}(x,t)$, and its corresponding stochastic
process $\sigma_{dyn}(X_{t},t)$. %
} the stochastic variable as $X$ and the state as $x$. In addition
to the drift $F(x,t)$, the process is driven by a feedback control
signal $u(x,t)\in\mathbb{R}^{m}$.

We call realizations of the above process \textquotedbl{}particles\textquotedbl{}.
Each particle is a trajectory that accumulates a state cost $V(x,t)$
and a quadratic control cost. This accumulated cost is called the
\textquotedbl{}path cost\textquotedbl{}. The aim is to find the control
function $u(x,t)$ that minimizes the expectation of the future path
cost with respect to the process (\ref{eq:controled process}). The
resulting optimal cost $J(x,t)$ at any time is called optimal cost-to-go,

\begin{equation}
J(x_{t},t)=\min_{u}\mathbb{E}_{u}\left[\int_{t}^{T}V(X_{s},s)+\frac{1}{2}||u(X_{s},s)||^{2}ds\right]\label{eq:optimal-cost-to-go}
\end{equation}
where the subscript $u$ denotes the feedback control function%
\footnote{To simplify the notation in a formula, we omit the arguments of functions
where the dependency is obvious from the context. Moreover, we some
times write $J(x_{t},t)$ to emphasize the dependency of a function
$J(x,t)$ on the momentary value $x_{t}$ of the trajectory $X_{[t,T]}$.%
} $u(x,s)$ for all $s\in[t,T]$ and $\left\Vert v\right\Vert ^{2}:=\sum_{i=1}^{m}v_{i}^{2}$
denotes the usual Euclidean norm squared for a vector $v\in\mathbb{R}^{m}$.
The expectation is defined as

\[
\mathbb{E}_{u}\left[R(X_{(t,T]})\right]:=\int dX_{(t,T]}p_{u}\left(X_{(t,T]}\right)R(X_{(t,T]})
\]
for any function $R(X_{(t,T]})$ of continuous trajectories starting
at a fixed $x_{t}$, $X_{(t,T]}:=(X_{s})_{s\in(t,T]\subset\mathbb{R}}|x_{t}$,
and $p_{u}\left(X_{(t,T]}\right):=p(X_{(t,T]}|x_{t},u)$. Notice that
this density is conditioned on the control function $u(x,s)$ for
all $s\in[t,T]$.

The optimal control 
\[
u^{*}(x_{t},t)=\mathrm{argmin}_{u}\mathbb{E}_{u}\left[\int_{t}^{T}V(X_{s},s)+\frac{1}{2}||u(X_{s},s)||^{2}ds\right]
\]
is the solution to this minimization.

We can express the expectation over the trajectories in (\ref{eq:optimal-cost-to-go})
as a Kullback-Leibler divergence between a distribution over trajectories
under the controlled dynamics (\ref{eq:controled process}) and the
uncontrolled dynamics (\ref{eq:Euler-Maruyama}). To see this consider
the following. In the limit of $ds\to0$, the transition density between
time $s$ and $s+ds$ for the controlled process is given by a Gaussian
\[
\hat{f}(x_{s+ds}|x_{s},u)=\mathcal{N}\left(x_{s+ds}|x_{s}+\tilde{F}ds,\sigma_{dyn}\sigma_{dyn}'\right),
\]
where $\tilde{F}=F(x_{s},s)+\sigma_{dyn}(x_{s},s)u(x_{s},s)$, $\sigma_{dyn}=\sigma_{dyn}(x_{s},s)$
and $'$ denotes transpose. This density is proportional to (see e.g.
\cite[Appendix B]{Kappen2012}) 
\begin{equation}
\hat{f}(x_{s+ds}|x_{s},u=0)\exp\left(\frac{1}{2}\left\Vert u(x_{s},s)\right\Vert ^{2}ds+u(x_{s},s)dW_{s}\right).\label{eq:change-of-measure-controlled-dynamics}
\end{equation}
Multiplying \eqref{eq:change-of-measure-controlled-dynamics} for
all times $s$ on the interval $(0,T]$, the distribution over controlled
dynamics is proportional to the distribution over the uncontrolled
dynamics (both conditioned on the initial state $x_{0}$) as 
\begin{multline}
p_{u}\left(X_{(0,T]}|x_{0}\right)=p_{0}\left(X_{(0,T]}|x_{0}\right)\times\dots\\
\exp\left(\frac{1}{2}\int_{0}^{T}\left\Vert u(X_{s},s)\right\Vert ^{2}ds+\int_{0}^{T}u(X_{s},s)dW_{s}\right).\label{eq:Grisanov-correction}
\end{multline}
From this, we derive that%
\footnote{Note that $\mathbb{E}_{u}\left[\int_{t}^{T}u(X_{s},s)dW_{s}\right]=0$
as a stochastic integral.%
} 
\[
\mathbb{E}_{u}\left[\log\frac{p_{u}\left(X_{(0,T]}|x_{0}\right)}{p_{0}\left(X_{(0,T]}|x_{0}\right)}\right]=\mathbb{E}_{u}\left[\int_{0}^{T}ds\frac{1}{2}||u(X_{s},s)||^{2}\right]
\]
is the KL divergence between the distribution over trajectories under
the control $u(x,t)$ and the distribution over trajectories under
the uncontrolled dynamics. Thus, the optimal cost-to-go at $t=0$
is 
\begin{equation}
J(x_{0})=\min_{u}\mathbb{E}_{u}\left[V(X_{[0,T]})+\log\frac{p_{u}\left(X_{(0,T]}|x_{0}\right)}{p_{0}\left(X_{(0,T]}|x_{0}\right)}\right]\label{eq:KL-diverg}
\end{equation}
where we define $V(X_{[0,T]}):=\int_{0}^{T}dsV(X_{s},s)$.

Since the feedback control function $u(x,t)$ determines fully the
distribution $p_{u}$, we can replace the minimization w.r.t. $u(x,t)$
with a minimization with respect to $p_{u}$ subject to the normalization
constraint $\int dX_{(0,T]}p_{u}\left(X_{(0,T]}|x_{0}\right)=1$.
The optimal control distribution conditioned on the initial state
$x_{0}$ that minimizes (\ref{eq:KL-diverg}) is then given by 
\begin{equation}
p_{u^{*}}(X_{(0,T]}|x_{0})=\frac{1}{\psi(x_{0})}p_{0}(X_{(0,T]}|x_{0})\exp\left(-V(X_{[0,T]})\right)\label{eq:opt-control-distrib}
\end{equation}
where the normalization constant is given by $\psi(x_{0}):=\mathbb{E}_{u=0}\left[\exp\left(-V(X_{[0,T]})\right)\right]$;
see \cite{Kappen2012} for details. If we identify $V(X_{[0,T]})=-\log\left[p(y_{0:T}|X_{[0,T]})\right]$
we see that the smoothing distribution for fixed initial state $x_{0}$
is identical to the optimal control distribution (\ref{eq:opt-control-distrib}):
\begin{equation}
p(X_{(0,T]}|y_{0:T},x_{0})=p_{u^{*}}(X_{(0,T]}|x_{0})\label{eq1}
\end{equation}
When the initial state $X_{0}$ is drawn from a prior distribution
$p_{0}(X_{0})$ the smoothing distribution (\ref{eq:smoothing-distribution})
is related to the optimal control distribution via 
\begin{equation}
p(X_{[0,T]}|y_{0:T})=p_{u^{*}}(X_{(0,T]}|X_{0})p(X_{0}|y_{0:T})\label{eq2}
\end{equation}
with 
\begin{equation}
p(X_{0}|y_{0:T})=\frac{\psi(X_{0})p_{0}(X_{0})}{p(y_{0:T})}\label{eq:initial-marginal}
\end{equation}
the posterior over the initial state.

Thus, we identify the problem of sampling from the joint smoothing
distribution with a stochastic control problem. We see that the optimal
control yields a distribution over trajectories that coincides exactly
with the smoothing distribution.

\section{Importance Sampling as Controlled Diffusion\label{sec:Method}}

In this section, we show how sampling from the posterior can be done
using controlled diffusions. We use a previous result that shows that
when the control approaches the optimal control, the quality of the
sampling, measured as the effective sample size, increases \cite{Thijssen2015}.
In general, we cannot compute the optimal control. We introduce the
APIS method that adapts feedback controllers to optimize the sampling
process.

\subsection{Importance sampling and the Relation to Optimal Control\label{sub:IS-and-PI}}

Eq. (\ref{eq:smoothing-distribution}) suggest that we can sample
from the smoothing distribution by sampling from the prior process
and weighting each trajectory $X_{[0,T]}$ with $p(y_{0:T}|X_{[0,T]})$.
We can use the control theory to improve the efficiency of the sampling.
Combining Eqs. (\ref{eq2}), (\ref{eq1}), (\ref{eq:opt-control-distrib})
and (\ref{eq:Grisanov-correction}) we can write 
\begin{multline}
p(X_{[0,T]}|y_{0:T})\propto p(X_{0}|y_{0:T})p_{u}(X_{(0,T]}|X_{0})\times\dots\\
\exp\left(\sum_{j=1}^{J}\log[g(y_{t_{j}}|X_{t_{j}})]-\int_{0}^{T}\frac{1}{2}||u_{s}||^{2}ds-\int_{0}^{T}u_{s}dW_{s}\right)\label{eq:importance-sampling-control}
\end{multline}
where we recall that $t_{J}=T$ and denote $u_{s}:=u(X_{s},s)$ for
simplicity. We can thus sample from $p_{u}$ and correct with the
exponential term, i.e. an importance sampling procedure for diffusion
processes \cite{kappen2016adaptive,Thijssen2015,Zhang2014,Fleming1982,Mitter1982,girsanov1960transforming}.
We call $u(x,t)$ the importance sampling control. In addition, we
use importance sampling with a proposal distribution $q(X_{0})$ to
sample from (\ref{eq:initial-marginal}).

We sample $i=1,\ldots,N$ particle trajectories. For each particle,
we define an importance weight 
\begin{equation}
\alpha_{u}=\exp[-S_{u}]\label{eq:weights}
\end{equation}
\[
S_{u}:=-\sum_{j=1}^{J}\log[g(y_{t_{j}}|X_{t_{j}})]+\int_{0}^{T}\frac{1}{2}||u_{s}||^{2}ds+\int_{0}^{T}u_{s}dW_{s}+S^{0}
\]
and $S^{0}:=-\log\left[\frac{p_{0}(X_{0})}{q(X_{0})}\right]$ and
normalize such that $\sum_{i=1}^{N}\alpha_{u}^{i}=1$. Notice that
the weights $\alpha_{u}$ depend on all observations $y_{0:T}$ and
on $u(x,t)$ through $S_{u}$.

The quality of the sampling can be quantified in terms of the effective
sample size, which we define as \cite{liu1996metropolized} 
\begin{equation}
ESS=\frac{N_{eff}}{N}=\frac{1}{Var(\alpha_{u})+1}.\label{eq:ESS}
\end{equation}
with $Var(\alpha_{u})$ the empirical variance in the $N$ sample
weights. We see that reducing the variance of the weights increases
the efficiency of the sampling procedure. In \cite{Thijssen2015}
upper and lower bounds for $Var(\alpha_{u})$ were found. The upper
bound 
\begin{equation}
Var(\alpha_{u})\leq\int_{0}^{T}\mathbb{E}_{u}\left\{ \left\Vert \alpha_{u}\left[u^{*}(X_{t},t)-u(X_{t},t)\right]\right\Vert ^{2}\right\} dt\label{eq:upper bound var}
\end{equation}
shows that the optimal control function $u^{*}(x,t)$ is the optimal
importance sampler in the sense that the importance weights have zero
variance and the ESS becomes maximal. Hence, the better we approximate
$u^{*}(x,t)$, the higher the efficiency of our importance sampler
will be.

\subsection{Adaptive Path Integral Smoother\label{sub:APIS}}

\begin{algorithm}
\caption{Adaptive Path Integral Smoother}

\begin{algorithmic}[1] 
\State \textbf{Input:} Observations $y_{0:T}$, prior $p_0(x_{0})$, control parametrization  $u^{0}(x,t)=A^{0}(t)h(x,t)$, learning rate $\eta<1$, particles $N$, iterations $I_{max}$, ESS threshold $\theta_{ess}\leq1$, annealing factor $\beta>1$ and annealing threshold $\gamma\geq0$. 
\State \textbf{Output:} Smoothing particle system $\{x_{[0,T]}^{i},\alpha_{u}^{i}\}_{i=1:N}$ and importance controller $u(x,t)$.
\State Set $n \leftarrow 0$
\While{$ESS<\theta_{ess}$ or $n\leq I_{max}$} 
\If {$n=0$}{ $x_{0}^{i}\sim p_0(x_{0}) $ for $i=1,\dots,N$}
\Else
\State $x_{0}^{i} \sim \mathcal{N}(\hat{\mu}_{0},\hat{\sigma}_{0}^{2}) $ for $i=1,\dots,N$
\State $S_{u}^i = -\log(p_0(x_{0}^i)/\mathcal{N}(x_{0}|\hat{\mu}_{0},\hat{\sigma}_{0}^{2}))$ 
\EndIf
\State Generate: $\{x_{[0,T]}^{i},\alpha_{u}^{i}\}_{i=1:N}$ according to (\ref{eq:controled process}) and (\ref{eq:weights}). 
\State Estimate $ESS$ from (\ref{eq:ESS})
\While{$ESS<\gamma$}
\State $S_{u}^i \leftarrow S_{u}^{i}/\beta$ for $i=1,\dots,N$
\State Estimate $\alpha_{u}$ from (\ref{eq:weights})
\State Estimate $ESS$ from (\ref{eq:ESS})
\EndWhile
\State Compute: $\hat{\mu}_{0},\hat{\sigma}_{0}^{2}$ from $\{x_{0}^{i},\alpha_{u}^{i}\}_{i=1:N}$ with (\ref{eq:adaptiveimportancesamplingforq}). 
\For{$t=0,\dots,T$} 
\State Estimate: $H(t)$ and $dQ(h_{t})$ with (\ref{eq:explicit-HandQ})
\State Update: $A_{t} \leftarrow A_{t}+\eta\frac{dQ(h_{t})}{dt}H(t)^{-1}$ 
\EndFor 
\State $n \leftarrow n+1$
\EndWhile
\end{algorithmic}
\end{algorithm}

Clearly, it is difficult to compute the optimal control in general.
However, we can efficiently estimate a suboptimal control using the
approach introduced in \cite{Thijssen2015}. Assume that the optimal
control can be approximately parametrized as 
\begin{equation}
u^{*}(x,t)=A^{*}(t)h(x,t)\in\mathbb{R}^{m}\label{eq:param cotrol}
\end{equation}
where $A^{*}(t)\in\mathbb{R}^{m\times k}$ are time-dependent parameters
and $h:\mathbb{R}^{n}\times\mathbb{R}\rightarrow\mathbb{R}^{k}$ are
the $k$ \textquotedbl{}basis\textquotedbl{} functions of the feedback
controller. In addition, we choose the importance sampling control
to be parametrized with the same basis functions: $u(x,t)=A(t)h(x,t)$.
The main theorem in \cite{Thijssen2015} implies 
\begin{equation}
A^{*}(t)\left\langle h_{t}\otimes h_{t}\right\rangle =A(t)\left\langle h_{t}\otimes h_{t}\right\rangle +\lim_{\delta t\rightarrow0}\frac{\left\langle \int_{t}^{t+\delta t}dW_{s}\otimes h_{s}\right\rangle }{\delta t}\label{eq:estm-opt-control}
\end{equation}
where $h_{t}:=h(X_{t},t)$, $h_{t}\otimes h_{t}$ is the outer product
$(h_{t}\otimes h_{t})_{kk'}=h_{k}(X_{t},t)h_{k'}(X_{t},t)$ and $\left\langle \bullet\right\rangle =\mathbb{E}_{u}[\alpha_{u}\bullet]$
is the weighted average targeting the posterior.

In practice the limit in the right side of (\ref{eq:estm-opt-control})
may lead to numerical instability when estimated with a finite number
of particles and time discretization $dt>0$. Therefore, one may consider
taking $\delta t\geq dt$ which yields a smoothed biased estimate
of $u(x,t)$ with less variance. Around observations, the control
may be a sensitive function of time and a small $\delta t$ is required.
In the reminder of the article we set $\delta t=dt$.

Equation (\ref{eq:estm-opt-control}) describes a procedure to compute
an estimate of the optimal control $u^{*}(x,t)$ based on an importance
sampling control $u(x,t)$. We can iterate this idea where in iteration
$r$ we estimate $A^{r}(t)$ as $A^{*}(t)$ in (\ref{eq:estm-opt-control})
with samples that we generate with a control function with parameters
$A^{r-1}(t)$ from the previous iteration. Then, (\ref{eq:estm-opt-control})
becomes 
\begin{equation}
A^{r+1}(t)=A^{r}(t)+\eta\frac{dQ(h_{t})}{dt}H^{-1}(t)\label{eq:update-rule}
\end{equation}
where $H(t)=\left\langle h_{t}\otimes h_{t}\right\rangle _{r}\in\mathbb{R}^{k\times k}$
and $dQ(h_{t}):=\left\langle dW_{t}\otimes h_{t}\right\rangle _{r}\in\mathbb{R}^{m\times k}$.
The learning rate $\eta<1$ accounts for sample errors at the beginning
of the learning procedure, when the ESS is low.

We need to estimate $H(t)$ and $dQ(h_{t})$. Both can be obtained
by sampling $N$ particles via numerical integration of (\ref{eq:controled process})
and weighting each with its corresponding $\alpha_{u}$. Then, the
expectation at each time $t$ is a weighted average over the particle
system $\{X_{t}^{i}\}_{i=1,\dots,N}$, 
\begin{multline}
H(t)=\frac{1}{N}\sum_{i=1}^{N}\alpha_{u}^{i}h(X_{t}^{i},t)\otimes h(X_{t}^{i},t)\\
dQ(h_{t})=\frac{1}{N}\sum_{i=1}^{N}\alpha_{u}^{i}dW_{t}^{i}\otimes h(X_{t}^{i},t)\label{eq:explicit-HandQ}
\end{multline}
where $dW_{t}^{i}$ is the noise realization of the $i$-th particle
at time $t$.

The posterior initial state $p(X_{0}|y_{0:T})$ is sampled using a
Gaussian adaptive importance sampling distribution $q(X_{0})=\mathcal{N}(X_{0}|\hat{\mu}_{0},\hat{\sigma}_{0}^{2})$
with $\hat{\mu}_{0}$ and $\hat{\sigma}_{0}^{2}$ the mean and covariance
of the marginal posterior at time $t=0$. After the first iteration,
we update 
\begin{equation}
\hat{\mu}_{0,l}=\left\langle X_{0,l}\right\rangle \quad\left(\hat{\sigma}_{0}^{2}\right)_{kl}=\left\langle (X_{0,k}-\hat{\mu}_{0,k})(X_{0,l}-\hat{\mu}_{0,l})\right\rangle \label{eq:adaptiveimportancesamplingforq}
\end{equation}

Equations (\ref{eq:adaptiveimportancesamplingforq}), (\ref{eq:update-rule})
and (\ref{eq:explicit-HandQ}) define the Adaptive Path Integral Smoother
(APIS) that learns iteratively the feedback controller defined in
(\ref{eq:param cotrol}). This is an adaptive importance sampling
procedure to obtain samples from the joint smoothing distribution
using controlled diffusion. Note that the control parameters $A(t)$
are estimated for each time $t$ independently.

The APIS algorithm starts by sampling from the uncontrolled dynamics.
We initialize the particles from $q(X_{0})=p_{0}(X_{0})$ if possible,
otherwise they are initialized from a proposal distribution $q(X_{0})$
of our choice. For subsequent iterations, the update rules (\ref{eq:adaptiveimportancesamplingforq}),
(\ref{eq:update-rule}) and (\ref{eq:explicit-HandQ}) are repeated
until the ESS reaches a threshold $\theta_{ess}\leq N_{eff}/N$ or
a maximal number of iterations $I=I_{max}$. The resulting weighted
particles give an estimate of the smoothing distribution. Alternatively,
one can check if the variance of the weights or the ESS has changed
significantly in the last $l$ iterations and stop if the change is
small.

The number of particles $N$ is one of the most important parameters.
The variance of the estimates reduces with $N$. In practice, we need
a large number of particles to ensure sufficiently good estimates.
The complexity of APIS is ${\cal O}(IN)$, where $I$ is the number
of adaptation iterations and $N$ the number of particles.

The learning rate $\eta$ determines the rate at which the control
function $u$ increases from its initial value zero. We observe poor
improvement of the importance sampler in terms of the ESS for large
learning rate $\eta$. In our experiments, we find good results with
$\eta\in[0.001,0.05]$ depending on the variance of the estimations.

Special attention is required at the initial iterations. Since the
initial importance sampler is very poor, the ESS is extremely low
and the estimates (\ref{eq:update-rule}) and (\ref{eq:explicit-HandQ})
are very inaccurate. For this reason we artificially increase the
ESS to a predetermined minimum number of particles $N_{0}$ by introducing
an 'adaptive annealing procedure' with a temperature $\lambda>1$
that scales the cost of each particle $i$ as $S_{u}^{i}\rightarrow S_{u}^{i}/\lambda$.
For a given set of particles we can then estimate the ESS for different
values of lambda $\lambda$. The smallest value of $\lambda$ such
that $ESS\approx N_{0}/N$ is found by setting $\lambda=\beta^{m}$
with $m=0,1,2,\ldots$ and $\beta>1$. The annealing factor $\beta$
should be chosen not too large to prevent overshoot, and not too small
to restrict the number of $m$ steps. We find that values of $\beta\in[1.05,1.15]$
prove to work well and finding $\lambda$ is very fast. The adaptive
annealing procedure is done whenever the ESS is below the threshold
$\gamma=N_{0}/N$. In our experiments we use $N_{0}=100-150$.

\section{Results\label{sec:Results}}

In this section, we present numerical results to show the efficiency
and accuracy of APIS compared to FS and FFBSi. Additionally, we show
the scaling of APIS to very high number of observations when the importance
control has the correct parametrization. 

For all numerical experiments, we fix the choice of the basis functions
to a linear feedback term and an open-loop controller (no state dependence,
only time dependence). For details on this choice and the implementation
we refer the reader to the appendix.

\subsection{Linear Quadratic System\label{sub:A-simple-system}}

\subsubsection{Low Likelihood Observation}

\begin{figure}[t]
\begin{centering}
\includegraphics[width=3.5in]{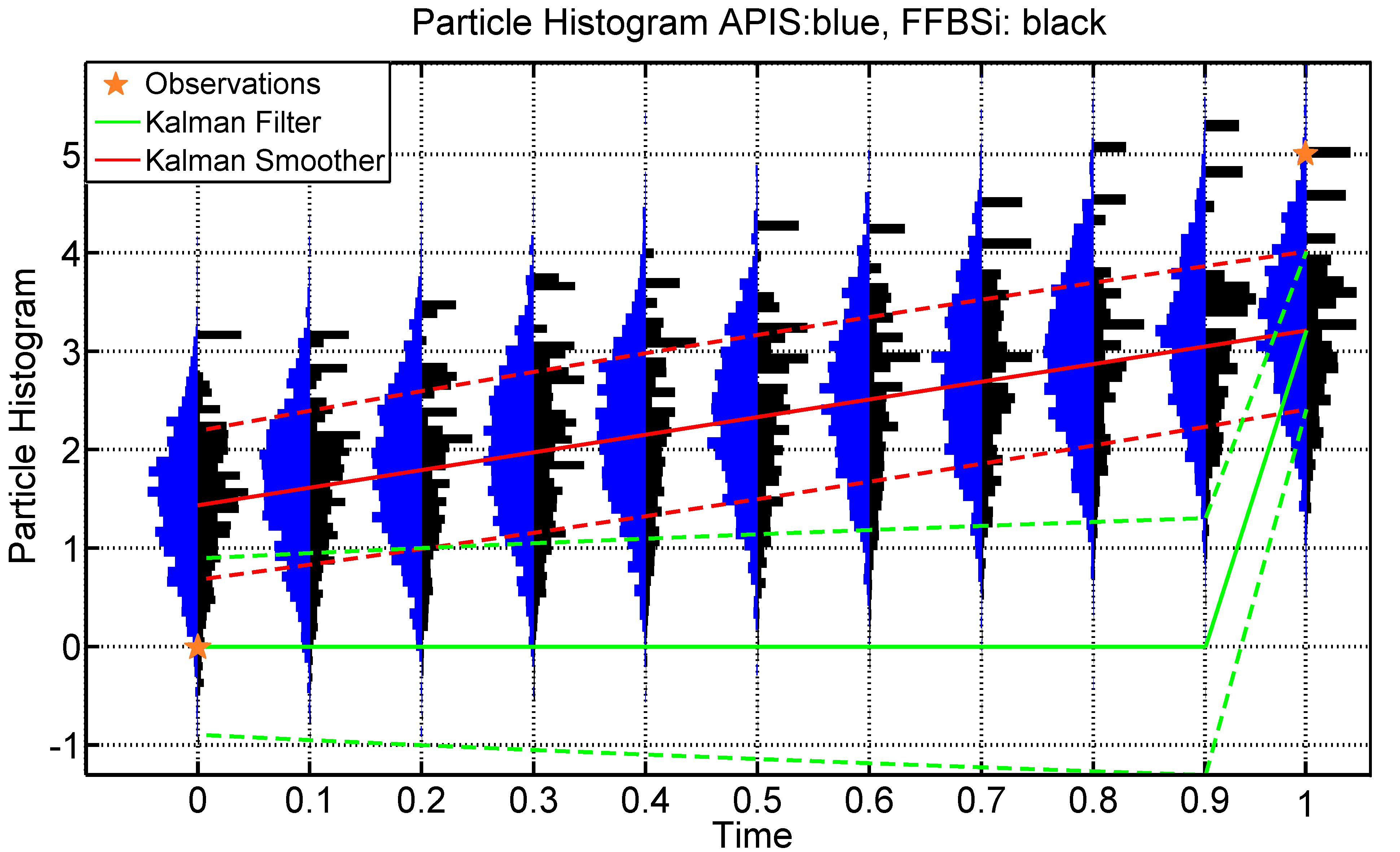}
\par\end{centering}

\caption{Kalman Smoother solution\label{fig:Kalman-Smoother-solution}. Notice
the small overlap of the filtering (green) and smoothing (red) solutions
due to the unlikely observation (orange) at $y_{T}=5$. Violin plots
(histograms) of particles obtained by FFBSi (black) and APIS (blue):
snapshots every $\Delta t=0.1$ starting at $t=0$. Notice the poor
particle representation in FFBSi. For APIS we used the following parameters:
$N=2000$ particles, learning rate $\eta=0.2$, no annealing procedure
and $I_{max}=15$ iterations. For FFBSi we used $N=M=2000$ forward
and backward particles. Color figures online.}
\end{figure}
\begin{figure}
\centering{}%
\begin{minipage}[t]{1\columnwidth}%
\begin{center}
\includegraphics[width=3.5in]{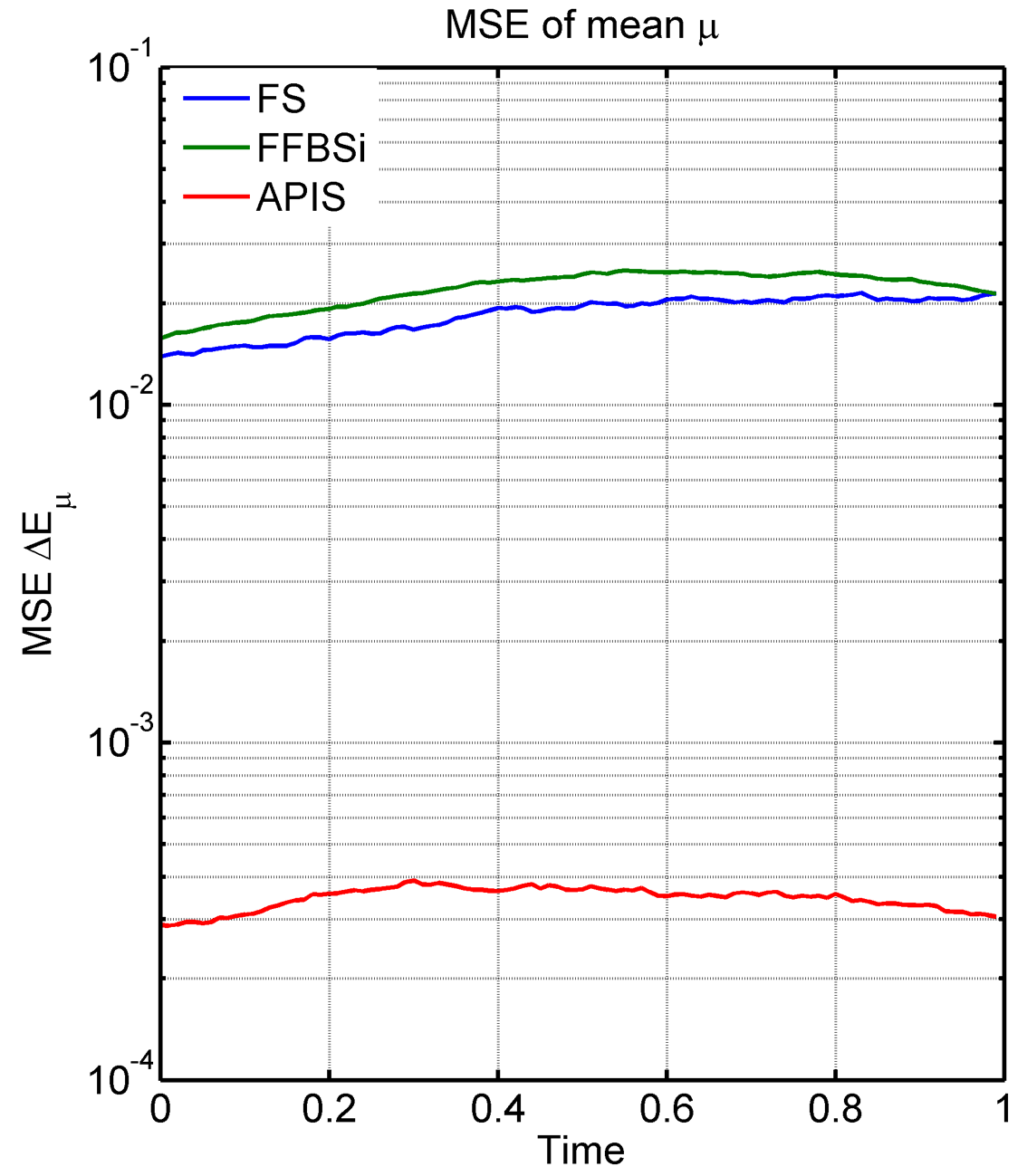}
\par\end{center}

\begin{center}
\caption{MSE of mean $\hat{\mu}$ over time. Estimates averaged over $R=250$
runs to avoid effects of the particular sampling realizations. We
used $N=2000$ forward particles in all methods and $M=2000$ backward
particles in FFBSi; observations at $y_{0}=0$ and $y_{T}=5$. Notice
the error for APIS (red) at all times; it is two orders of magnitude
lower than FS and FFBSi. In APIS, estimations are made using the last
particle system obtained after $I_{max}=15$ iterations and without
annealing ($\gamma=0$).\label{fig:MSE-of-mean}}

\par\end{center}%
\end{minipage}\hfill{}%
\begin{minipage}[t]{1\columnwidth}%
\begin{center}
\includegraphics[width=3.5in]{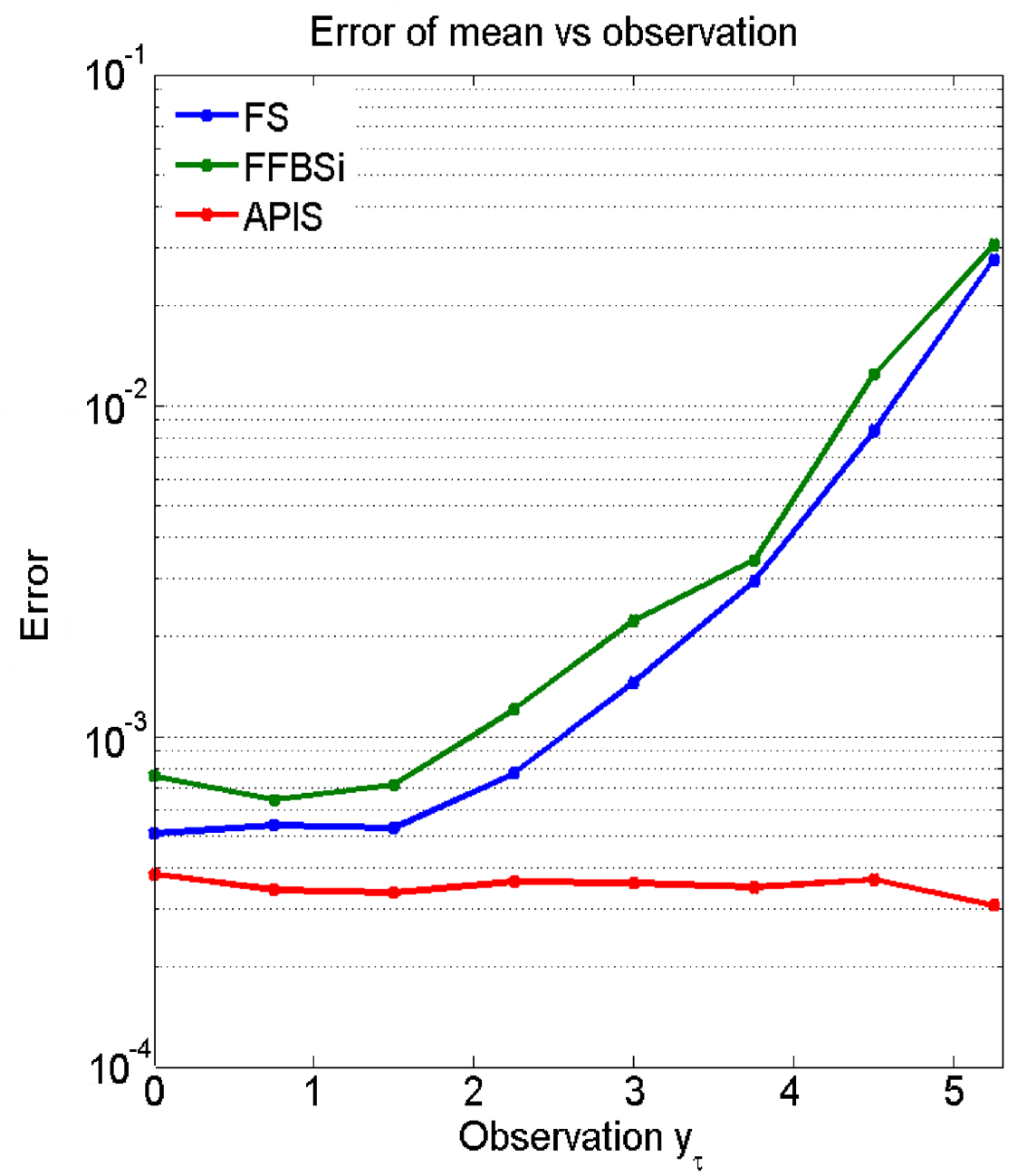}\caption{Error $\hat{E}_{\mu}$ for $y_{T}\in\{0,0.75,1.5,2.25,3,3.75,4.5,5.25\}$
and always $y_{0}=0$: For each $y_{T}$, we estimate $\widehat{\Delta E}_{\mu}(t)$
using 100 runs and $N=M=2000$ particles. Notice the logarithmic scale
in the error-axis. The error in the variance is similar. APIS has
no annealing in this example. \label{fig:BM-Time-averaged-MSE}}

\par\end{center}%
\end{minipage}
\end{figure}

Consider a Brownian motion $X_{s+ds}\sim\mathcal{N}(x_{s},\sigma_{dyn}^{2}dt)$
with $\sigma_{dyn}=1,dt=0.01$ and a Gaussian observation model $y_{t}\sim\mathcal{N}(x_{t},\sigma_{obs}^{2}=1)$
for $t=0,T$. We fix the observations at $y_{0}=0$, $y_{T}=5$ and
the length of the series at $T=1$. The initial distribution $p_{0}(X_{0})$
is a Gaussian centered at $x_{0}=0$ with variance $\sigma_{0}^{2}=4$.
The exact solution for this model is given by the Kalman smoother,
Figure \ref{fig:Kalman-Smoother-solution}. Notice the poor overlap
of the filtering and smoothing distributions. 

We compare the particle smoothing distribution given by APIS and FFBSi.
In Figure \ref{fig:Kalman-Smoother-solution}, we show violin plots%
\footnote{We used distributionPlot.m for MATLAB%
} for a particular realization of the particles at times $t=0,0.1,0.2,\dots,1$.
Although $N$ is large, FFBSi poorly represents the Gaussian posterior
marginal distributions. The effect worsens for large $t\lesssim T$,
where filtered and smoothed marginals differ most. The histograms
for the bootstrap filter-smoother (FS) are similar to those of FFBSi
(not shown). On the contrary, APIS histograms represent much better
the Gaussian distribution. 

If the filtered particles do not represent the smoothing distribution
well enough, the backward pass will have a low ESS and therefore the
backward particles will mix poorly. Although we observe an increase
in the averaged ESS of the backward pass from $2\%$ at $t\lesssim T$
to $7\%$ for times $0\lesssim t$, this is not enough to improve
the estimations, see Figure \ref{fig:MSE-of-mean}. For comparison,
APIS increases the ESS of the whole path from 1.5\% to 98\% in 15
iterations by adapting the trajectories from the initial filtering
distribution to the smoothing distribution. 

We can use the exact solution to compare the performance of all methods
using the mean squared error (MSE)
\[
\widehat{\Delta E}_{\mu}(t)=\frac{1}{R}\sum_{j=1}^{R}(\hat{\mu}_{j}(t)-\mu_{KS}(t))^{2}
\]
 where $\mu_{KS}$ is the mean of the ground truth obtained by the
Kalman smoother, $\hat{\mu}_{j}$ is the estimated mean of each method
in run $j$ and $R$ the number of runs. In each run, we used the
same parameters as above. Figure \ref{fig:MSE-of-mean} shows $\widehat{\Delta E}_{\mu}$
versus $t$. We observe that APIS has an accuracy two orders of magnitude
higher than FS and FFBSi. The errors of the variance estimates are
very similar to the errors of the mean estimates. 

Note the slight increase of the MSE of FFBSi vis-�-vis FS. This may
be due to the small number of observations. In this example, there
is no gain in applying a backward pass. For larger time series we
observe an improvement of the estimates in FFBSi compared to FS, as
is to be expected. However, APIS was consistently better in all the
examples studied.

Figure \ref{fig:BM-Time-averaged-MSE} shows the error of the mean
$\widehat{E}_{\mu}=\frac{1}{T}\int_{0}^{T}\widehat{\Delta E}_{\mu}(s)ds$
as a function of the unlikely observation $y_{T}$. Notice how the
performance of both FS and FFBSi are comparable to APIS if the observation
is close to the high density region of the filtering ($y_{T}\in[0,2]$)
but deteriorates very fast for unlikely observations. On the contrary,
the error in APIS is virtually independent of the position $y_{T}$
of the observation. This is due the adaptation of APIS to the likelihood.

\subsubsection{High Number of Observations}

\begin{figure*}[t]
\centering{}%
\begin{minipage}[t]{1\columnwidth}%
\begin{center}
\includegraphics[width=3.5in]{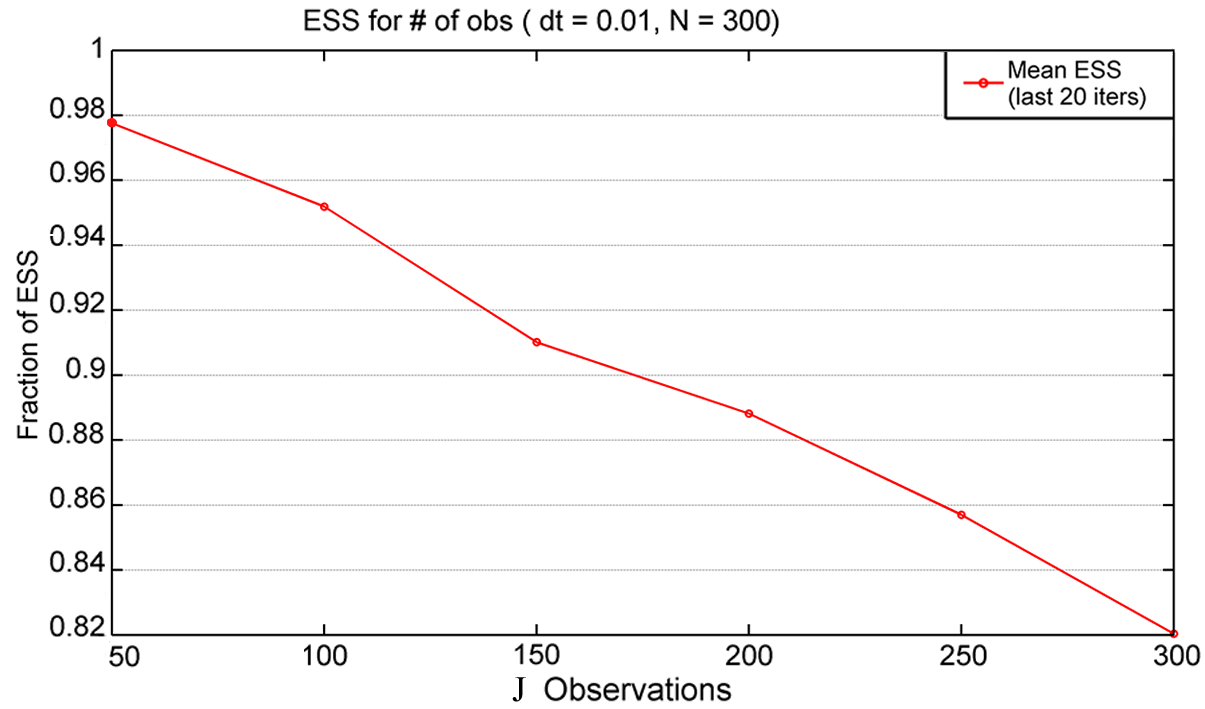}
\par\end{center}

\caption{ESS estimated for 100, 200 and 300 observations. We use a learning
rate of $\eta=0.05$, $N=300$ particles, $I_{max}=100$ and no annealing
($\gamma=0$). The ESS decays slowly with the number of observations.\label{fig:ESS_vs_Nobs}}
\end{minipage}\hfill{}%
\begin{minipage}[t]{1\columnwidth}%
\begin{center}
\includegraphics[width=3.5in]{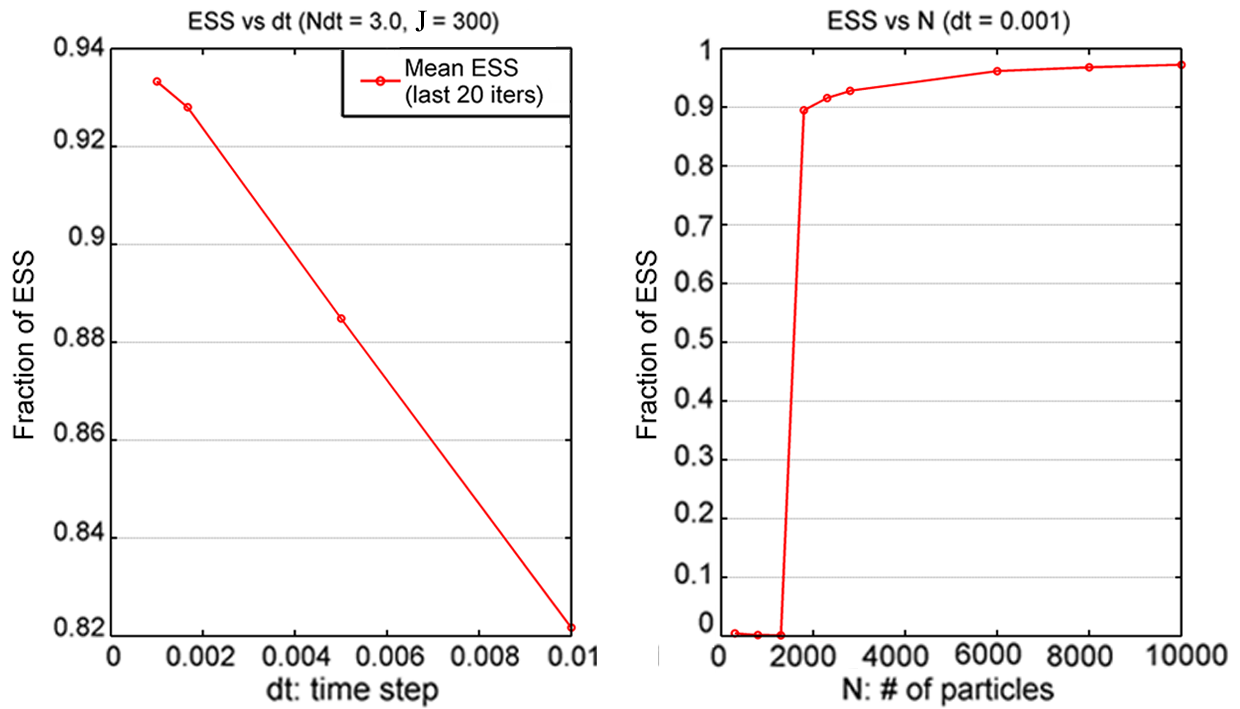}
\par\end{center}

\caption{Mean ESS over the last 20 iterations for a time series of $300$ observations.
Left: ESS for increasing precision. While $dt$ decreases, $N$ increases
such that $Ndt=3$. Right: ESS for increasing number of particles.
The integration step is fixed at $dt=10^{-3}$. The APIS parameters
$\eta,I_{max}$ and $\gamma$ are the same as in Figure \ref{fig:ESS_vs_Nobs}.\label{fig:ESS-vs-N}}
\end{minipage}
\end{figure*}

Now, we study in more detail the ESS of APIS for higher number of
observations $J$. Consider again a Brownian motion with $\sigma_{dyn}^{2}=0.75,\sigma_{obs}^{2}=0.9$
and a time horizon $T=3$. The prior $p_{0}(X_{0})$ is as before.
We generate a single time series of $300$ observations on the time
interval $[0,T]$ and define a posterior estimation problem given
the first $100$, $200$ and all observations. For each problem we
estimate the ESS after 100 iterations. 

In Figure \ref{fig:ESS_vs_Nobs} we observe a slow decay of the ESS
when the number of observations $J$ increases. This decay can be
compensated with an increase in the precision of the estimations.
We illustrate this by increasing the number of samples $N$ while
decreasing the integration step $dt$ such that $Ndt$ remains constant.
The consequence is an increase of the ESS for $dt\rightarrow0$ as
in Figure \ref{fig:ESS-vs-N} left. On the right of the same figure,
we show the ESS for the same time series of 300 observations but with
fixed $dt$ and incrementing $N$. We observe a fast increase of the
ESS and saturation%
\footnote{Accordingly, the MSE of both estimators decreased very fast until
it saturated at a much lower value due to estimation errors in the
controller (not shown).%
} for higher $N$. 

The observed ESS in \prettyref{fig:ESS-vs-N} can be improved even
further by simply decreasing the learning rate $\eta$. Naturally,
learning the controller will take longer, thus, more iterations are
needed. However, for the same number of particles $N=1000$ the ESS
reached with $\eta=0.01$ is about 83\% compared to 69\% in \prettyref{fig:ESS-vs-N},
where we set $\eta=0.05$.  

The excellent scaling with the number of observations is due to the
correct parametrization of the importance control function. However,
for small samples sizes $N$, the variance is too large to efficiently
bootstrap APIS and an increase in ESS is not guaranteed. The minimum
amount of particles needed to bootstrap APIS is problem dependent.
However for a given problem, the discretization step $dt$ has a big
impact on the choice of $N$. As a rule-of-thumb, we find that for
a fixed $dt$, one must choose at least $N>2/dt$ to have stable results.

More complex problems require higher number of samples $N$. In this
case, the annealing procedure helps to avoid prohibitive large $N$.
We considered now $J=1000$ for the same system and parameters as
above except that we anneal the weights if the ESS is below a threshold
$\gamma=0.01$ ($\beta=1.15$). This allows us to use only $N=10^{4}$
particles. Above $\gamma$ there is no annealing anymore and the raw
ESS converges to a value around $0.6$, which is an increase of 3
orders of magnitude vis-�-vis the uncontrolled dynamics. After learning,
the absolute error of the mean $\left|\hat{\mu}_{APIS}-\mu_{KS}\right|$
stays lower than $0.01$ over time and the averaged absolute error
is $1.8\times10^{-3}$. The absolute error in the variance is similar.

We can bootstrap APIS because we obtain a higher ESS from the annealed
particle system. This allows us to estimate a control that improves
the raw ESS incrementally. Without annealing, the ESS stays at $2\times10^{-4}$
even after 1000 iterations. This result shows the importance of the
adaptive annealing procedure when the number of observations is very
large or the problem too complex.

The above analysis shows that the ESS in APIS scales very well with
the number of observations given the correct parametrization of the
controller. Moreover, the error of the estimates stays small over
the whole time interval.

\subsection{A Neural Network Model\label{sub:A-Neural-Network}}

We consider a non-linear system and examine the performance of APIS,
FFBSi and FS. In this example the linear feedback control is clearly
suboptimal. However, we show that the variance of the estimates is
lower for APIS than for FFBSi and FS.

We consider a 5 dimensional non-linear neural network described by
\[
dX_{t}=-X_{t}dt+\tanh(BX_{t}+\theta+A\sin(\omega t))dt+\sigma_{dyn}dW_{t}
\]
where $B\in\mathbb{R}^{5\times5}$ and $\theta\in\mathbb{R}^{5}$
are an antisymmetric connectivity matrix and a threshold vector respectively.
The elements of the vector $A\in\mathbb{R}^{5}$ are the amplitudes
of independent sinusoidal inputs with frequencies given by $\omega\in\mathbb{R}^{5}$.
We choose the values randomly from Gaussian distributions $\theta_{i}\backsim\mathcal{N}(0,\sigma_{\theta}=0.75),\ A_{i}\backsim\mathcal{N}(0,\sigma_{A}=2),\ \omega_{i}\backsim\mathcal{N}(\pi/5,\sigma_{\omega}=\pi)$
and $B_{ij}\backsim\mathcal{N}(0,\sigma_{B}=2)$ with $B_{ij}=-B_{ji}$,
for all $i,j=1,\dots,5$. In addition, we set $\sigma_{dyn}^{2}=0.05$
and an integration step of $dt=0.01$. 

Furthermore, we assume a Gaussian observation model with $Y_{t_{i}}\backsim\mathcal{N}(X_{1}(t_{i}),\sigma_{obs}=0.1)$
for $i=1,\dots,J$ and sample an observation every $\Delta_{obs}=10dt$.
Note that only one of the five neural states is observed.

\begin{figure}[t]
\begin{centering}
\includegraphics[width=3.5in]{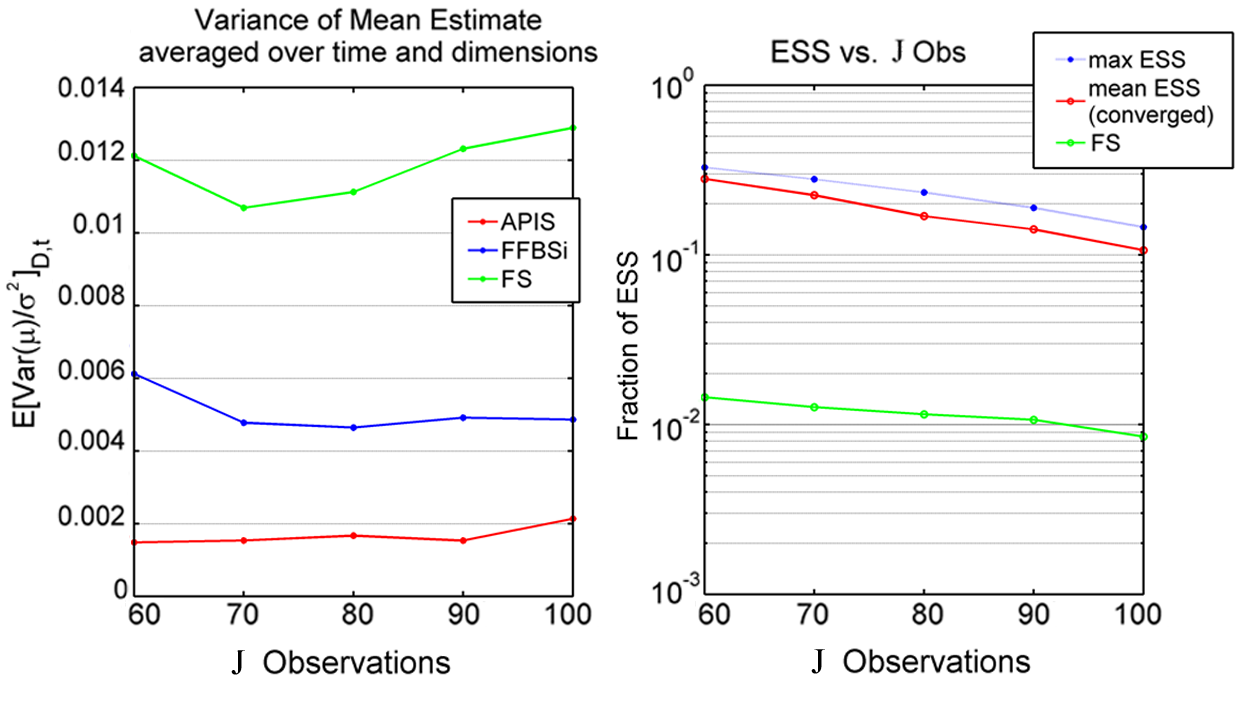}
\par\end{centering}

\caption{Left: variance of the mean $\hat{\mu}$ across 10 estimations, scaled
by the variance of the posterior. This is averaged over time and dimensions
to give a single measure for each $J$. Right: Effective Sample Size
(ESS) for different number of observations $J$. Notice the logarithmic
scale. The ESS of FS is taken as the number of unique trajectories.
We use the same amount of forward particles $N=6000$ in all 3 methods,
and the number of backward particles is set such that the CPU time
spend on FFBSi and APIS is similar. The estimation of the posterior
in APIS is accepted when a predefined threshold of $\theta_{ess}=0.1$
is reached. Each algorithm was repeated $R=10$ times to estimate
the variance. We used a fixed initial condition $x_{0}=0$ and $\gamma=0.02$
in APIS. \label{fig:NN-ESS_vs_Nobs}}
\end{figure}

We consider now a fixed initial condition at $x_{0}=0$ to examine
the ESS of APIS and FS without the effect of importance sampling in
the initial state. In Figure \ref{fig:NN-ESS_vs_Nobs} right, we compare
the ESS of APIS and FS as a function of the number of observations%
\footnote{It is computationally challenging to compute the ESS of the $M$ backward
trajectories in FFBSi, so we do not consider the ESS of FFBSi.%
}. The ESS is an order of magnitude higher for APIS than for FS, but
the efficiency of both decrease with the number of observations. The
ESS of APIS starts at around $30\%$ for 60 observations and ends
at around $10\%$ for 100 observations. Moreover, the ESS cannot be
increased much further with higher precision. This is the result of
a suboptimal importance control. 

Nevertheless, the efficiency and performance of APIS clearly increases
compared to FS. As seen in Figure \ref{fig:NN-ESS_vs_Nobs} left,
the variance of $\hat{\mu}$ in APIS is significantly lower than in
FS and FFBSi. We have similar results for the variance of $\hat{\sigma}^{2}$. 

\begin{figure}[t]
\centering{}\includegraphics[clip,width=3.5in]{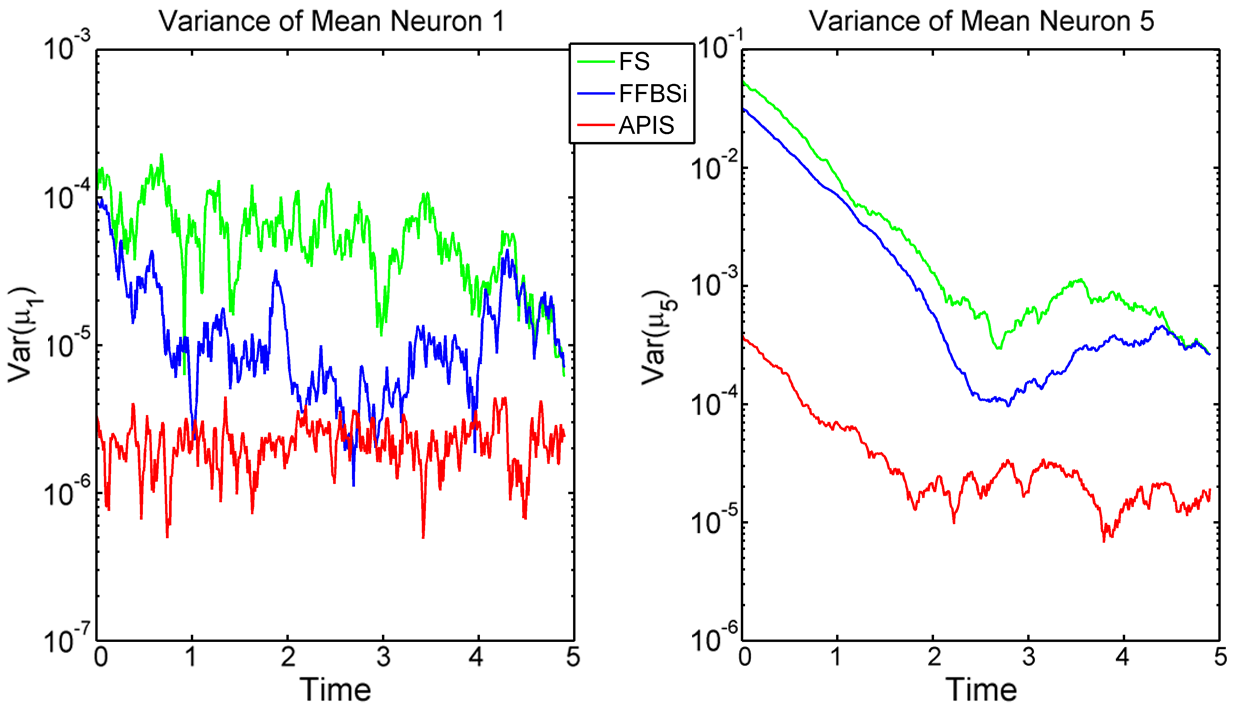}\caption{Variance of mean estimate $\hat{\mu}$ for the partially observed
neuron 1 (left) and for the hidden neuron 5 (right). The variance
is obtained from $R=12$ estimates. Green: FS; blue: FFBSi; red: APIS.
Notice that APIS has a lower variance up to two orders of magnitude
(log-scale). The estimations for all other neurons 2,3,4 are similar
to neuron 5. The setting is similar as in Figure \ref{fig:NN-ESS_vs_Nobs}
but with $J=50$. In APIS, we use $\gamma=0.02$ and $N=7500$ forward
particles. The ESS threshold is set to $\theta_{ess}=0.2$. In FS
and FFBSi we use $N=5000$ forward particles and $M=2500$ backward
particles such that APIS and FFBSi have again the same CPU time available.\label{fig:VarEstim-Neuron1and5} }
\end{figure}

Now, consider a Gaussian prior $p_{0}(X_{0})$ with mean $\mu_{0}=0$
and variance $\sigma_{0}^{2}=1$. We study the performance of the
three methods. Figure \ref{fig:VarEstim-Neuron1and5} left shows the
variance of the mean for the observed neuron. All three methods have
reliable estimates but both FS and FFBSi have higher variance. APIS
keeps the variance of the estimates consistently lower over the entire
time interval. 

Figure \ref{fig:VarEstim-Neuron1and5} right shows that the variance
of the mean for the hidden neurons is up to two orders of magnitude
higher for FS and FFBSi than for APIS. The increased variance towards
earlier times is in part an effect of the importance sampling procedure
at the initialization, which affects all methods. Nevertheless, the
adaptation of the proposal distribution $q(X_{0})$ in APIS reduces
significantly this effect.

\begin{figure*}
\begin{centering}
\includegraphics[width=7.15in]{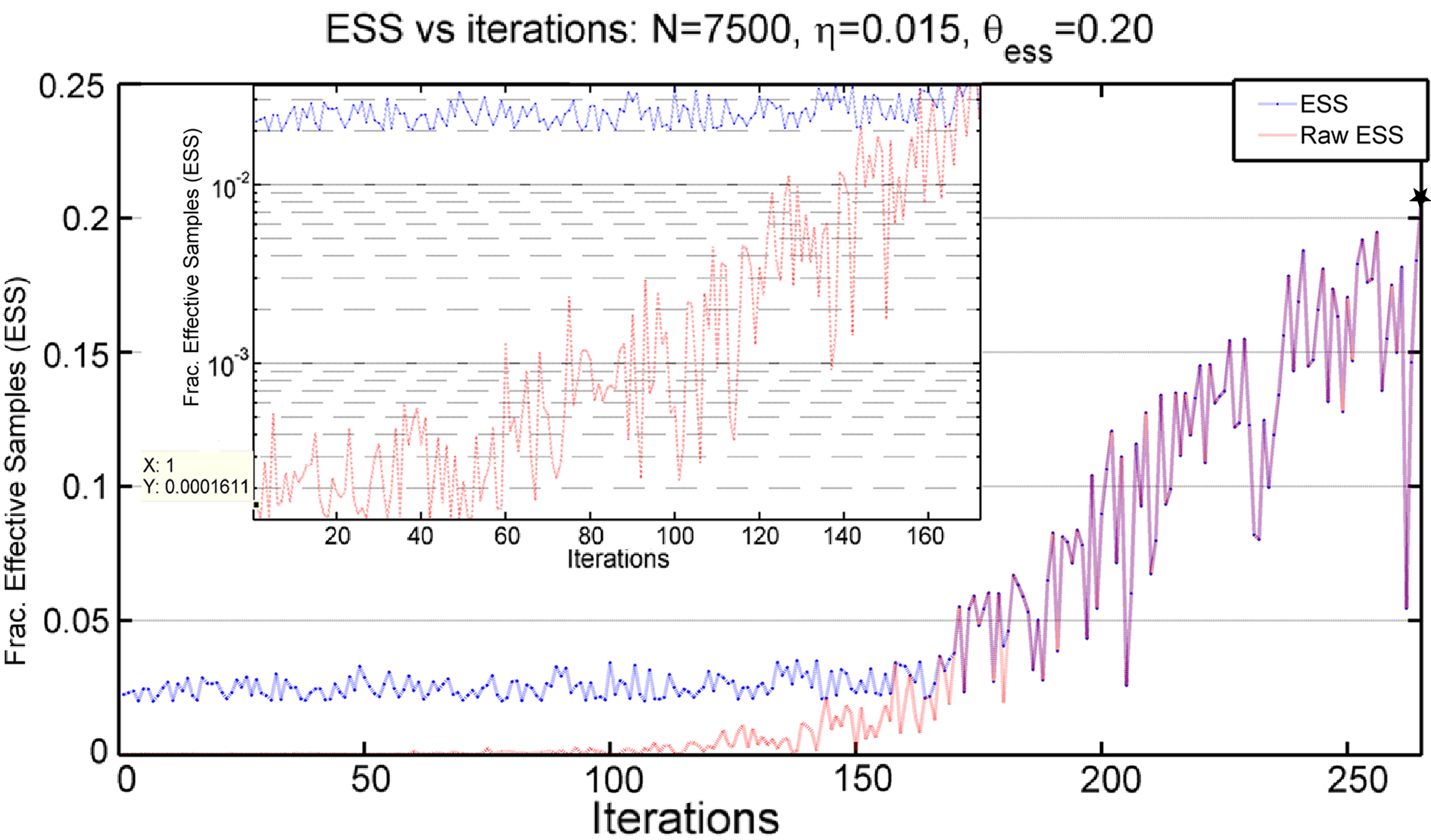}
\par\end{centering}

\caption{Effective Sample Size (ESS) for a single run. Black marker symbolizes
accepted samples used for the estimation of the smoothing distribution
($\theta_{ess}=0.2$).\label{fig:NN-ESS_vs_Iters}}
\end{figure*}

Finally in Figure \ref{fig:NN-ESS_vs_Iters}, we show the typical
improvement of the ESS for this example. At the beginning, the ESS
is around $2\%$ due to annealing (blue line, $\lambda>1$) and the
raw ESS increases from $0.02\%$ up to $2\%$ (red dotted line, $\lambda=1$).
This increase is due to the control estimations obtained from the
annealed particle system. After the ESS surpasses $\gamma$, APIS
reaches the stopping threshold $\theta_{ess}$ in about 80 iterations.
The final ESS of $20\%$ is a remarkable improvement vis-�-vis the
ESS of the posterior marginals in FS, which stays most of the time
below $10\%$ and around $2\%$ for times close to $t=0$.

\section{Discussion\label{sec:Discussion}}

In this work, we present a new smoothing algorithm for diffusion processes
in continuous time. This method estimates iteratively a feedback controller
to target the posterior distribution. We show that having the correct
parametrization of the control, we can sample the posterior with very
high efficiency and observe excellent scaling with the number of observations.
Furthermore, even with a suboptimal controller for a non-linear system
the ESS increases by several orders of magnitude and the variance
of the estimates is up to two orders of magnitude lower than the variance
of FS and FFBSi.

We are aware of many important developments in particle methods, some
of them having a linear computational complexity in the number of
particles, e.g. \cite{Fearnhead2010,Dubarry2011,Bunch2013}. However,
we use the standard FFBSi to show the degeneracy of the backward pass
and compare the results with APIS. We think that a comprehensive comparison
of all state-of-the-art methods goes beyond the scope of this work
but deserves to be addressed in the future.

More efficient proposal distributions are used in practice for FFBSi,
e.g. by linearization of the discretized SDE \cite{Doucet2000}. However,
in general this is only valid for sufficiently short time intervals,
which might be shorter than the interval between observations and
could lead to errors in the integration of \prettyref{eq:Euler-Maruyama}
and the estimations. Also, similar schemes can be used to improve
the efficiency of APIS.

The optimal resampling step at time $t-1$ in the ${\cal O}(N)$ two-filter
algorithm \cite{Fearnhead2010} is given by marginals of the form
$p(y_{t:T}|X_{t-1})w_{t-1}$ where $w_{t-1}$ are the filter weights.
This is approximated by a single observation \textquotedbl{}look-ahead\textquotedbl{}
distribution $p(y_{t}|X_{t-1})w_{t-1}$. It is interesting to notice
that APIS effectively implements a \textquotedbl{}look-ahead\textquotedbl{}
transition probability considering all observations. This makes APIS
an attractive alternative to the ${\cal O}(N)$ two-filter algorithm.

One can apply the ideas of this paper also to discrete state, discrete
time problems. A discussion on discrete systems in \cite{todorov2006linearly,Kappen2012}
shows that we can frame discrete HMMs as a control problem. One can
then adapt the uncontrolled dynamics using the Cross Entropy idea
\cite{de2005tutorial}. In addition, the application to continuous
time hidden Markov jump processes is also interesting and possible.
Both problems are equivalent to the optimization of a KL divergence
similar to (\ref{eq:KL-diverg}) \cite{opper2008variational}. But
the details are different and one would need to work out the details
on how a control theoretic approach would influence the process rates
to perform importance sampling. Furthermore, details on the learning
procedure applied to both types of systems and the parametrization
of the feedback controller need to be worked out.

It is interesting to notice the similarities between the iterated
auxiliary particle filter%
\footnote{We are grateful to a reviewer for pointing out this work and the similarity
between APIS and this method.%
} \cite{guarniero2015iterated} and optimal control solution in \cite{todorov2006linearly,Kappen2012}.
The computation of the optimal twisted functions in \cite{guarniero2015iterated}
is related to the backward message passing involved in the computation
of the optimal transition probability. However, the connection to
optimal control was not pointed out in this work. Recall that, in
practice, the functions and transition densities in \cite{guarniero2015iterated}
have to be restricted. On the contrary, APIS has great flexibility
in the design of the controller. Hence, it may prove fruitful to explore
the similarities between both approaches to develop better approximation
schemes for the iterative auxiliary particle methods.

Similarly, it is interesting to contrast the ideas of this paper to
those in \cite{cappe2008adaptive,cornebise2008adaptive}, where the
aim was to minimize the KL divergence between the target density and
a parametrized (mixture) model. These ideas are similar in flavor
to the ideas here%
\footnote{We are grateful to a reviewer for mentioning this approach and its
possible relation to APIS for discrete HMMs.%
} but it is not obvious how both schemes relate. The difficulty in
the comparison resides in the components learned in each method. While
we aim at learning a parametrized controller to adapt the prior (uncontrolled)
dynamics, \cite{cornebise2008adaptive} directly modifies the adjustment
multiplier weights and the proposal kernels. These modifications may
correspond to the importance control correction term and the introduction
of a control term in the dynamics. However, a detailed comparison
may help us to understand further the relation between adaptive importance
sampling and control for general hidden state processes.

In our experiments we have initialized the importance sampler with
$u(x,t)=0$. One can consider better initializations of the controller
using other methods. For instance, one could initialize a linear feedback
controller around the solution to the optimal trajectory as in \cite{pequito2011nonlinear}.
The initialization of the controller will have an impact on the performance
that is not to be taken lightly, for instance, bad importance control
$u$ might decrease the ESS and result in poor estimates which could
lead to a further decrease in the quality of the controller. Our experience
is, however, that a linear feedback controller initialized with the
uncontrolled dynamics is a robust procedure when combined with annealing.

Naturally, the initialization of the particles at $t=0$ has an impact
on the ESS. In this paper we have chosen (axis-aligned) Gaussians
as proposal distributions to target the posterior marginal at $t=0$,
but other initializations are possible. For instance, one can sample
from a multivariate Gaussian with general covariance matrix, a general
kernel density estimator to deal with multi-modal distributions or
initialize the particles via the Cross Entropy method \cite{de2005tutorial}.

The choice of the number of particles $N$ and the annealing threshold
$\gamma$ is an open question. On the one hand, we know that increasing
$N$ influences the efficiency of APIS in a non-linear way and that
a value below some threshold prevents APIS from bootstrapping. However,
it is not obvious how to choose $N$ in an efficient way. On the other
hand, with the annealing procedure introduced in this paper, it is
possible to bootstrap APIS without increasing $N$ to prohibitive
sizes. Thus, it is important to understand how both, the number of
particles and the annealing procedure, influence the learning of the
control.

Unfortunately, there is no proof that APIS converges to the optimal
control within the class of control solutions constrained by the parametrization.
In practice we can use the ESS as a quality measure, which can be
used to asses the \textquotedbl{}goodness\textquotedbl{} of the chosen
parametrization. Still, the question of optimality given a parametrization
is a very interesting question that deserves to be explored.

The efficiency of APIS depends on the choice of basis functions, which
is problem dependent. This choice is an open question. The linear
basis functions that we considered in this paper are very robust to
learn, however, they might lead to problems whenever the posterior
has multiple pronounced modes. Since the update rule (\ref{eq:update-rule})
poses no restriction on the basis functions, more complex functions
are possible. Thus, in problems with multiple modes, locally linear
functions such as $x\exp(-x^{2})$ may be a good choice. 

More general strategies to estimate an efficient importance control
are possible. For instance, we can use a nested set of functions.
During the initial iterations a simple function (say linear) is learned
and as soon as the ESS is sufficiently high, non-linear extensions
are learned. In this context the application of universal function
approximations such as deep neural networks may be promising.

\section{Acknowledgment}

H.-Ch. Ruiz would like to thank Sep Thijssen, Dominik Thalmeier and
George S. Stamatescu for helpful discussions. He would also like to
show his gratitude to Silvia Menchon and the reviewers for their useful
comments on an earlier version of the manuscript.

\appendix[Implementation Details]{\label{Implementation-Details}}

The proposed APIS method was discussed in \prettyref{sub:APIS} in
general form to remark that in principle any linear parametrization
of the controller can be learned. Here, we discuss the implementation
details for the results in \prettyref{sec:Results}. 

We use a linear feedback controller standardized w.r.t. the target
distribution, i.e. $h(x,t):=(1,z(x,t))'$, where $z(x,t),x\in\mathbb{R}^{m}$.
Each component $z_{i}$ is defined as 
\[
z_{i}(x,t)=\frac{x_{i}-\mu_{i}(t)}{\sigma_{i}(t)}
\]
where $\mu_{i}(t)=\left\langle X_{t,i}\right\rangle ,\sigma_{i}^{2}(t)=\left\langle (X_{t,i}-\mu_{i}(t))^{2}\right\rangle ;i=1,\dots,m$
are the mean and variance of the state components w.r.t. smoothing
marginal at time $t$. The values are initialized in the first iteration
as $\mu_{i}(t)=0$ and $\sigma_{i}^{2}(t)=1$ for all times. This
choice of basis functions splits (\ref{eq:update-rule}) such that
the updates for the open-loop and feedback controllers are independent
and numerically more stable.

For completeness, we give the explicit update rules for the standardized
linear feedback controller. The control has a very simple form $u(x,t)=a(t)z(x_{t},t)+b(t)$,
where $a(t)\in\mathbb{R}^{m\times m}$ is a square matrix of the same
dimension as the state and $b(t)\in\mathbb{R}^{m}$ is an open-loop
controller. Then, the cross-correlation matrix becomes 
\[
H(t)=\left[\begin{array}{cc}
1 & 0\\
0 & C(t)
\end{array}\right]
\]
where $C(t)$ is the correlation matrix of the state variables. We
have component-wise, 
\[
C^{ij}(t)=\left\langle \frac{[X_{t,i}-\mu_{i}(t)][X_{t,j}-\mu_{j}(t)]}{\sigma_{i}(t)\sigma_{j}(t)}\right\rangle _{r}.
\]

For $dQ_{r}(h_{t})$ we have a matrix in $\mathbb{R}^{m\times(m+1)}$
with elements 
\begin{gather*}
[dQ_{r}(1)]^{i1}=\left\langle dW_{t,i}\right\rangle _{r}\\
{}[dQ_{r}(z_{t})]^{i(j+1)}=\left\langle dW_{t,i}\frac{X_{t,j}-\mu_{j}(t)}{\sigma_{j}(t)}\right\rangle _{r}
\end{gather*}
for each $i,j=1,\dots,m$.

This gives the explicit update rules, 
\begin{gather*}
b^{r+1}(t)=b^{r}(t)+\eta\frac{\left\langle dW_{t}\right\rangle _{r}}{dt}\\
a^{r+1}(t)=a^{r}(t)+\eta\frac{dQ_{r}(z_{t})}{dt}C^{-1}(t)
\end{gather*}

We use as prior distribution $p_{0}(X_{0})$ for the state at time
$t=0$ a Gaussian with mean and variance $\mu_{0},\sigma_{0}^{2}$,
respectively. In addition, we use an adaptive Gaussian as proposal
distribution $q(X_{0})$. Thus, we initialize at each iteration of
APIS the cost to correct for this importance sampling step. This initial
value is given for particles $l=1,\dots,N$ by 
\[
S^{0,l}=\sum_{i=1}^{m}\frac{(X_{0,i}^{l}-\hat{\mu}_{0,i})^{2}}{2\hat{\sigma}_{0,i}^{2}}-\frac{(X_{0,i}^{l}-\mu_{0,i})^{2}}{2\sigma_{0,i}^{2}}
\]
where $\hat{\mu}_{0}=\mu(0)$ and $\hat{\sigma}_{0,i}^{2}=\sigma_{i}(0)^{2}$
are the mean and variance of the posterior marginal at $t=0$.

The control cost over the entire interval $[0,T]$ is given by the
linear feedback controller and approximated by the discretization
step $dt$, 
\begin{gather*}
\int_{0}^{T}\frac{1}{2}\left\Vert u(X_{s},s)\right\Vert ^{2}ds+\int_{0}^{T}u(X_{s},s)'dW_{s}\approx\\
\sum_{i=1}^{L}\left\Vert a(t_{i})z_{t_{i}}+b(t_{i})\right\Vert ^{2}\frac{dt}{2}+\left(a(t_{i})z_{t_{i}}+b(t_{i})\right)'dW_{t_{i}}
\end{gather*}
where $z_{t_{i}}=z(X_{t_{i}},t_{i})$ and $dW_{t_{i}}$ the noise
realization at time $t_{i}$ with variance $\sigma^{2}=dt$. The summation
goes over the $L=T/dt$ integration steps.

In general, there is a trade-off between the amount of iterations
and the number of particles needed, but we observe that the convergence
of the ESS to a maximal value is very fast once it has bootstrapped,
so usually a number of iterations $I_{max}\ll N$ can be chosen and--as
a rule of thumb--higher $N$ allows for less iterations. The particles
can be sampled independently so this step is parallelizable. Nevertheless,
when reducing the learning rate, it is possible to reduce by at least
one order of magnitude the number of particles needed to bootstrap
APIS. Naturally, this will increase the number of iterations needed.
In addition, we found that an annealing procedure with $\gamma$ in
the range $0.02-0.05$ works well for low $N$.

Finally, for FFBSi and FS, we use the algorithms as described in \cite[Algorithm 4]{Lindsten2013}
with the numerical integration of the SDE (\ref{eq:Euler-Maruyama})
as proposal distribution for the filtering. We initialized particles
according to $p_{0}(X_{0})$ unless noted otherwise.

\bibliographystyle{IEEEtran}
\bibliography{REFS_APIS_IEEE_bibtex}

\begin{thebibliography}{10}
\providecommand{\url}[1]{#1}
\csname url@samestyle\endcsname
\providecommand{\newblock}{\relax}
\providecommand{\bibinfo}[2]{#2}
\providecommand{\BIBentrySTDinterwordspacing}{\spaceskip=0pt\relax}
\providecommand{\BIBentryALTinterwordstretchfactor}{4}
\providecommand{\BIBentryALTinterwordspacing}{\spaceskip=\fontdimen2\font plus
\BIBentryALTinterwordstretchfactor\fontdimen3\font minus
  \fontdimen4\font\relax}
\providecommand{\BIBforeignlanguage}[2]{{%
\expandafter\ifx\csname l@#1\endcsname\relax
\typeout{** WARNING: IEEEtran.bst: No hyphenation pattern has been}%
\typeout{** loaded for the language `#1'. Using the pattern for}%
\typeout{** the default language instead.}%
\else
\language=\csname l@#1\endcsname
\fi
#2}}
\providecommand{\BIBdecl}{\relax}
\BIBdecl

\bibitem{JulierUhlmann1997}
S.~Julier and J.~Uhlmann, ``A new extension of the kalman filter to nonlinear
  systems,'' in \emph{Int. Symp. Aerospace/Defense Sensing, Simul. and
  Controls}, 1997.

\bibitem{sarkka2007unscented}
S.~Sarkka, ``On unscented kalman filtering for state estimation of
  continuous-time nonlinear systems,'' \emph{IEEE Transactions on automatic
  control}, vol.~52, no.~9, pp. 1631--1641, 2007.

\bibitem{sarkka2008unscented}
S.~S{\"a}rkk{\"a}, ``Unscented rauch--tung--striebel smoother,'' \emph{IEEE
  Transactions on Automatic Control}, vol.~53, no.~3, pp. 845--849, 2008.

\bibitem{sutter2015variational}
T.~Sutter, A.~Ganguly, and H.~Koeppl, ``A variational approach to path
  estimation and parameter inference of hidden diffusion processes,''
  \emph{arXiv preprint arXiv:1508.00506}, 2015.

\bibitem{kitagawa1996monte}
G.~Kitagawa, ``Monte carlo filter and smoother for non-gaussian nonlinear state
  space models,'' \emph{Journal of computational and graphical statistics},
  vol.~5, no.~1, pp. 1--25, 1996.

\bibitem{Godsill2004}
\BIBentryALTinterwordspacing
S.~J. Godsill, A.~Doucet, and M.~West, ``Monte carlo smoothing for nonlinear
  time series,'' \emph{J. Amer. Statist. Assoc.}, vol.~99, no. 465, pp.
  156--168, Mar 2004. [Online]. Available:
  \url{http://www.tandfonline.com/doi/abs/10.1198/016214504000000151}
\BIBentrySTDinterwordspacing

\bibitem{Doucet2000}
A.~Doucet, S.~Godsill, and C.~Andrieu, ``On sequential monte carlo sampling
  methods for bayesian filtering,'' \emph{Statist. Comput.}, vol.~10, no.~3,
  pp. 197--208, Jul 2000.

\bibitem{Lindsten2013}
\BIBentryALTinterwordspacing
F.~Lindsten, ``Backward simulation methods for monte carlo statistical
  inference,'' \emph{Foundations and Trends® in Machine Learning}, vol.~6,
  no.~1, pp. 1--143, 2013. [Online]. Available:
  \url{http://www.nowpublishers.com/articles/foundations-and-trends-in-machine-learning/MAL-045}
\BIBentrySTDinterwordspacing

\bibitem{bresler1986two}
Y.~Bresler, ``Two-filter formulae for discrete-time non-linear bayesian
  smoothing,'' \emph{International Journal of Control}, vol.~43, no.~2, pp.
  629--641, 1986.

\bibitem{Briers2010}
\BIBentryALTinterwordspacing
M.~Briers, A.~Doucet, and S.~Maskell, ``Smoothing algorithms for state-space
  models,'' \emph{Annals of the Institute of Statistical Mathematics}, vol.~62,
  no.~1, pp. 61--89, Feb 2010. [Online]. Available:
  \url{http://link.springer.com/10.1007/s10463-009-0236-2}
\BIBentrySTDinterwordspacing

\bibitem{Fearnhead2010}
\BIBentryALTinterwordspacing
P.~Fearnhead, D.~Wyncoll, and J.~Tawn, ``A sequential smoothing algorithm with
  linear computational cost,'' \emph{Biometrika}, vol.~97, no.~2, pp. 447--464,
  Jun 2010. [Online]. Available:
  \url{http://biomet.oxfordjournals.org/cgi/doi/10.1093/biomet/asq013}
\BIBentrySTDinterwordspacing

\bibitem{Doucet2009}
A.~M. Doucet, Arnaud \&~Johansen, ``A tutorial on particle filtering and
  smoothing: Fifteen years later,'' \emph{Handbook of Nonlinear Filtering},
  vol.~12, pp. 656--704, 2009.

\bibitem{liu2008monte}
J.~S. Liu, \emph{Monte Carlo strategies in scientific computing}.\hskip 1em
  plus 0.5em minus 0.4em\relax Springer Science \& Business Media, 2008.

\bibitem{fearnhead2008computational}
P.~Fearnhead, ``Computational methods for complex stochastic systems: a review
  of some alternatives to mcmc,'' \emph{Statistics and Computing}, vol.~18,
  no.~2, pp. 151--171, 2008.

\bibitem{del2012adaptive}
P.~Del~Moral, A.~Doucet, A.~Jasra \emph{et~al.}, ``On adaptive resampling
  strategies for sequential monte carlo methods,'' \emph{Bernoulli}, vol.~18,
  no.~1, pp. 252--278, 2012.

\bibitem{Chopin2004}
N.~Chopin, ``Central limit theorem for sequential monte carlo methods and its
  application to bayesian inference,'' \emph{Ann. Stat.}, vol.~32, no.~6, pp.
  2385--2411, 2004.

\bibitem{Douc2011}
\BIBentryALTinterwordspacing
R.~Douc, A.~Garivier, E.~Moulines, and J.~Olsson, ``Sequential monte carlo
  smoothing for general state space hidden markov models,'' \emph{The Annals of
  Applied Probability}, vol.~21, no.~6, pp. 2109--2145, Dec 2011. [Online].
  Available: \url{http://projecteuclid.org/euclid.aoap/1322057317}
\BIBentrySTDinterwordspacing

\bibitem{Kloeden2012}
P.~E. Kloeden, E.~Platen, and H.~Schurz, \emph{Numerical solution of SDE
  through computer experiments}.\hskip 1em plus 0.5em minus 0.4em\relax
  Springer-Verlag Berlin Heidelberg, 2012.

\bibitem{Murray2011}
\BIBentryALTinterwordspacing
L.~Murray and A.~Storkey, ``Particle smoothing in continuous time: A fast
  approach via density estimation,'' \emph{IEEE Transactions on Signal
  Processing}, vol.~59, no.~3, pp. 1017--1026, Mar 2011. [Online]. Available:
  \url{http://ieeexplore.ieee.org/lpdocs/epic03/wrapper.htm?arnumber=5654660}
\BIBentrySTDinterwordspacing

\bibitem{Taghavi2012}
\BIBentryALTinterwordspacing
E.~Taghavi, ``A study of linear complexity particle filter smoothers,''
  Master's thesis, Chalmers University of Technology, 2012. [Online].
  Available:
  \url{http://publications.lib.chalmers.se/records/fulltext/156741.pdf}
\BIBentrySTDinterwordspacing

\bibitem{Dubarry2011}
\BIBentryALTinterwordspacing
C.~Dubarry and R.~Douc, ``Improving particle approximations of the joint
  smoothing distribution with linear computational cost.''\hskip 1em plus 0.5em
  minus 0.4em\relax IEEE, Jun 2011, pp. 209--212. [Online]. Available:
  \url{http://ieeexplore.ieee.org/lpdocs/epic03/wrapper.htm?arnumber=5967661}
\BIBentrySTDinterwordspacing

\bibitem{Bunch2013}
\BIBentryALTinterwordspacing
P.~Bunch and S.~Godsill, ``Improved particle approximations to the joint
  smoothing distribution using markov chain monte carlo,'' \emph{IEEE
  Transactions on Signal Processing}, vol.~61, no.~4, pp. 956--963, Feb 2013.
  [Online]. Available:
  \url{http://ieeexplore.ieee.org/lpdocs/epic03/wrapper.htm?arnumber=6359869}
\BIBentrySTDinterwordspacing

\bibitem{guarniero2015iterated}
P.~Guarniero, A.~M. Johansen, and A.~Lee, ``The iterated auxiliary particle
  filter,'' \emph{arXiv preprint arXiv:1511.06286}, 2015.

\bibitem{de2005tutorial}
P.-T. De~Boer, D.~P. Kroese, S.~Mannor, and R.~Y. Rubinstein, ``A tutorial on
  the cross-entropy method,'' \emph{Annals of operations research}, vol. 134,
  no.~1, pp. 19--67, 2005.

\bibitem{cappe2008adaptive}
O.~Capp{\'e}, R.~Douc, A.~Guillin, J.-M. Marin, and C.~P. Robert, ``Adaptive
  importance sampling in general mixture classes,'' \emph{Statistics and
  Computing}, vol.~18, no.~4, pp. 447--459, 2008.

\bibitem{cornebise2008adaptive}
J.~Cornebise, {\'E}.~Moulines, and J.~Olsson, ``Adaptive methods for sequential
  importance sampling with application to state space models,''
  \emph{Statistics and Computing}, vol.~18, no.~4, pp. 461--480, 2008.

\bibitem{pitt1999filtering}
M.~K. Pitt and N.~Shephard, ``Filtering via simulation: Auxiliary particle
  filters,'' \emph{Journal of the American statistical association}, vol.~94,
  no. 446, pp. 590--599, 1999.

\bibitem{Pardoux1981}
\BIBentryALTinterwordspacing
E.~Pardoux, ``The solution of the nonlinear filtering equation as a likelihood
  function.''\hskip 1em plus 0.5em minus 0.4em\relax IEEE, Dec 1981, pp.
  316--319. [Online]. Available:
  \url{http://ieeexplore.ieee.org/lpdocs/epic03/wrapper.htm?arnumber=4046946}
\BIBentrySTDinterwordspacing

\bibitem{Fleming1982}
S.~K. Fleming, Wendell H \&~Mitter, ``Optimal control and nonlinear filtering
  for nondegenerate diffusion processes,'' \emph{Stochastics: An International
  Journal of Probability and Stochastic Processes}, vol.~8, no.~1, pp. 63--77,
  1982.

\bibitem{Mitter1982}
\BIBentryALTinterwordspacing
S.~K. Mitter, \emph{Nonlinear filtering of diffusion processes a guided tour},
  ser. Lecture Notes in Control and Information Sciences.\hskip 1em plus 0.5em
  minus 0.4em\relax Springer-Verlag, 1982, vol.~42, ch. chapter 23, pp.
  256--266. [Online]. Available:
  \url{http://www.springerlink.com/index/10.1007/BFb0004544}
\BIBentrySTDinterwordspacing

\bibitem{Kappen2005}
\BIBentryALTinterwordspacing
H.~J. Kappen, ``Linear theory for control of nonlinear stochastic systems,''
  \emph{Physical Review Letters}, vol.~95, no.~20, Nov 2005. [Online].
  Available: \url{http://link.aps.org/doi/10.1103/PhysRevLett.95.200201}
\BIBentrySTDinterwordspacing

\bibitem{yang2013feedback}
T.~Yang, P.~G. Mehta, and S.~P. Meyn, ``Feedback particle filter,''
  \emph{Automatic Control, IEEE Transactions on}, vol.~58, no.~10, pp.
  2465--2480, 2013.

\bibitem{pequito2011nonlinear}
S.~Pequito, P.~Aguiar, B.~Sinopoli, and D.~Gomes, ``Nonlinear estimation using
  mean field games,'' in \emph{NetGCOOP 2011: International conference on
  NETwork Games, COntrol and OPtimization}.\hskip 1em plus 0.5em minus
  0.4em\relax IEEE, 2011.

\bibitem{Thijssen2015}
\BIBentryALTinterwordspacing
S.~Thijssen and H.~J. Kappen, ``Path integral control and state-dependent
  feedback,'' \emph{Physical Review E}, vol.~91, no.~3, Mar 2015. [Online].
  Available: \url{http://link.aps.org/doi/10.1103/PhysRevE.91.032104}
\BIBentrySTDinterwordspacing

\bibitem{kappen2016adaptive}
H.~J. Kappen and H.~C. Ruiz, ``Adaptive importance sampling for control and
  inference,'' \emph{Journal of Statistical Physics}, vol. 162, no.~5, pp.
  1244--1266, 2016.

\bibitem{Kappen2011}
\BIBentryALTinterwordspacing
H.~J. Kappen, ``Optimal control theory and the linear bellman equation,''
  \emph{Inference and Learning in Dynamic Models}, pp. 363--387, 2011.
  [Online]. Available: \url{http://hdl.handle.net/2066/94184}
\BIBentrySTDinterwordspacing

\bibitem{Kappen2012}
\BIBentryALTinterwordspacing
H.~Kappen, V.~G\'omez, and M.~pper, ``Optimal control as a graphical model
  inference problem,'' \emph{Machine Learning}, vol.~87, no.~2, pp. 159--182,
  May 2012. [Online]. Available:
  \url{http://link.springer.com/10.1007/s10994-012-5278-7}
\BIBentrySTDinterwordspacing

\bibitem{Zhang2014}
\BIBentryALTinterwordspacing
W.~Zhang, H.~Wang, C.~Hartmann, M.~Weber, and C.~Schuette, ``Applications of
  the cross-entropy method to importance sampling and optimal control of
  diffusions,'' \emph{SIAM Journal on Scientific Computing}, vol.~36, no.~6,
  pp. A2654--A2672, Jan 2014. [Online]. Available:
  \url{http://epubs.siam.org/doi/abs/10.1137/14096493X}
\BIBentrySTDinterwordspacing

\bibitem{girsanov1960transforming}
I.~V. Girsanov, ``On transforming a certain class of stochastic processes by
  absolutely continuous substitution of measures,'' \emph{Theory of Probability
  \& Its Applications}, vol.~5, no.~3, pp. 285--301, 1960.

\bibitem{liu1996metropolized}
J.~S. Liu, ``Metropolized independent sampling with comparisons to rejection
  sampling and importance sampling,'' \emph{Statistics and Computing}, vol.~6,
  no.~2, pp. 113--119, 1996.

\bibitem{todorov2006linearly}
E.~Todorov, ``Linearly-solvable markov decision problems,'' in \emph{Advances
  in neural information processing systems}, 2006, pp. 1369--1376.

\bibitem{opper2008variational}
M.~Opper and G.~Sanguinetti, ``Variational inference for markov jump
  processes,'' in \emph{Advances in Neural Information Processing Systems},
  2008, pp. 1105--1112.

\end{thebibliography}

\begin{IEEEbiographynophoto}{Hans-Christian Ruiz}
was born in Mexico City in 1984. He received the degree Diplom-Physiker in theoretical physics from the University of Munich in 2011. In 2012, he worked on generalizations of spin networks during an internship at the Max-Planck Institute for Gravitational Physics in Potsdam, Germany. He is currently working as a PhD student at the Donders Institute in Nijmegen, the Netherlands. His work is on stochastic optimal control, Bayesian inference and neural networks. He has a Marie Curie fellowship of the Neural Engineering Transformative Technologies Initial Training Network.
\end{IEEEbiographynophoto}
\begin{IEEEbiographynophoto}{Bert Kappen}
 completed the PhD degree in particle physics in 1987 at the Rockefeller University, New York, USA. From 1987 until 1989 he worked as a scientist at the Philips Research Laboratories in Eindhoven, the Netherlands. Since 1989, he is conducting research on neural networks at the laboratory for biophysics of the University of Nijmegen, the Netherlands. Since 1997 he is associate professor and since 2004 full professor at this university. His group is involved in research on Bayesian machine learning, stochastic control theory, computational neuroscience and several applications in collaboration with industry. His research was awarded in 1997 the prestigious national PIONIER research subsidy. He co-founded in 1998 the company Smart Research, which sell prediction software based on neural networks. He has developed a medical diagnostic expert system called Promedas, which assists doctors to make accurate diagnosis of patients. He is director of the Dutch Foundation for Neural Networks (SNN), which coordinates research on neural networks and machine learning in the Netherlands. Since 2009, he is honorary faculty at UCL's Gatsby Computational Neuroscience Unit in London, UK.
\end{IEEEbiographynophoto}
\enlargethispage{-60mm}
\end{document}